\documentclass{article}

\usepackage{amsmath,amssymb,pifont}

\usepackage{multicol}
\usepackage{multirow}
\usepackage{amstext}
\usepackage{amsthm}
\usepackage{booktabs}
\usepackage[skip=0pt]{subcaption}
\usepackage{times}
\usepackage{lipsum}
\usepackage[shortlabels]{enumitem}
\usepackage{cancel}
\usepackage{wrapfig}
\usepackage{array}
\usepackage{siunitx}
\usepackage{csvsimple}
\usepackage[multidot]{grffile}
\usepackage{bbm}
\usepackage{float}
\usepackage{dblfloatfix}
\usepackage{hyperref}
\usepackage{makecell}
\usepackage{bbm, dsfont}
\usepackage{mathtools}
\usepackage[dvipsnames]{xcolor}
\usepackage{comment}
\usepackage[capitalise]{cleveref}

\usepackage{geometry}
\usepackage{enumitem}

\usepackage[multiple]{footmisc}
\usepackage{mathrsfs}
\usepackage{todonotes}
\usepackage{tikz}
\usepackage{hyperref}
\usepackage{cleveref}
\usepackage{algorithm}
\usepackage[noend]{algpseudocode}
\usepackage{xfrac}

\usepackage{graphicx} 
\usepackage{color}

\usepackage{array}
\usepackage{amssymb}
\usepackage{amsmath}
\usepackage{xspace}
\usepackage{fancyhdr}
\usepackage{comment}

\usepackage{tcolorbox}

\definecolor{darkgreen}{rgb}{0,0.4,0.0}

\newcommand{\inv}[1]{{{#1}^{-1}}}
\newcommand{\Cinv}{\inv{\bfC}}
\newcommand{\clipnorm}{\zeta}

\newcommand{\mdim}{m}  
\newcommand{\bmu}{\boldsymbol{\mu}}
\newcommand{\bnu}{\boldsymbol{\nu}}
\newcommand{\bluetext}[1]{{\color{blue}#1}}

\hypersetup{final}

\newcommand{\boldzero}{\ensuremath{\boldsymbol{0}}}
\newcommand{\boldone}{\ensuremath{\boldsymbol{1}}}

\newcommand{\bfC}{\ensuremath{\mathbf{C}}}
\newcommand{\bfD}{\ensuremath{\mathbf{D}}}

\newcommand{\bfI}{\ensuremath{\mathbf{I}}}

\newcommand{\bfa}{\ensuremath{\mathbf{a}}}

\newcommand{\bfg}{\ensuremath{\mathbf{g}}}

\newcommand{\bfs}{\ensuremath{\mathbf{s}}}

\newcommand{\bfu}{\ensuremath{\mathbf{u}}}
\newcommand{\bfv}{\ensuremath{\mathbf{v}}}

\newcommand{\bfz}{\ensuremath{\mathbf{z}}}

\newcommand{\calA}{\ensuremath{\mathcal{A}}}

\renewcommand{\Pr}{\mathop{\mathbf{Pr}}}

\newcommand{\R}{\mathbb{R}}

\makeatletter
\newcommand{\vast}{\bBigg@{4}}
\newcommand{\Vast}{\bBigg@{5}}
\makeatother

\newcommand{\ex}[2]{{\ifx&#1& \mathbb{E} \else
\underset{#1}{\mathbb{E}} \fi \left[#2\right]}}
\newcommand{\pr}[2]{{\ifx&#1& \mathbb{P} \else
\underset{#1}{\mathbb{P}} \fi \left[#2\right]}}

\newcommand{\Var}[1]{\ensuremath{\mathbf{Var}\left(#1\right)}}
\newcommand{\Cov}[2]{\ensuremath{\mathbf{Cov}\left(#1, #2\right)}}

\newcommand{\ltwo}[1]{\left\|#1\right\|_2}

\usepackage{ulem}

\DeclarePairedDelimiterX{\infdivx}[2]{(}{)}{%
  #1\;\delimsize\|\;#2%
}

\newcommand{\mypar}[1]{\smallskip
	\noindent{\textbf{{#1}:}}}
	
\renewcommand{\epsilon}{\varepsilon}

\renewcommand{\tilde}{\widetilde}

\newcommand{\clipop}{\operatorname{clip}}
\newcommand{\clip}[1]{\clipop\!\left( #1; \clipnorm \right)}

\newcommand{\bftheta}{\boldsymbol{\theta}}

\newcommand{\adagrad}{\textsc{AdaGrad}\xspace}
\newcommand{\dpadam}{\textsc{DP-Adam}\xspace}
\newcommand{\adam}{\textsc{Adam}\xspace}
\newcommand{\sgd}{\textsc{SGD}\xspace}
\newcommand{\dpsgd}{\textsc{DP-SGD}\xspace}
\newcommand{\dpmf}{\textsc{DP-MF}\xspace}
\newcommand{\bandmf}{\textsc{DP-BandMF}\xspace}
\newcommand{\stability}{\epsilon_s}

\DeclareMathOperator{\Tr}{Tr}
\DeclareMathOperator{\diag}{diag}
\DeclareMathOperator{\sign}{sign}

\setlist{nolistsep}
\setlist[itemize]{noitemsep, topsep=0pt}

\setlist{nolistsep}
\setlist[itemize]{noitemsep, topsep=0pt}

\usepackage{fullpage}
\usepackage[numbers, sort, comma, square]{natbib}

\begin{document}

\title{On Design Principles for Private Adaptive Optimizers}

\author{
Arun Ganesh\thanks{Google Research, \texttt{\{arunganesh, mcmahan\}@google.com}} \and
Brendan McMahan\footnotemark[1] \and 
Abhradeep Thakurta\thanks{Google DeepMind, \texttt{athakurta@google.com}}
}
\date{}
\maketitle

\begin{abstract}
The spherical noise added to gradients in differentially private (DP) training  undermines the performance  of adaptive optimizers like \adagrad and \adam, and hence many recent works have proposed algorithms to address this challenge. However, the empirical results in these works focus on simple tasks and models and the conclusions may not generalize to model training in practice. In this paper we survey several of these variants, and develop better theoretical intuition for them as well as perform empirical studies comparing them. We find that a common intuition of aiming for unbiased estimates of second moments of gradients in adaptive optimizers is misguided, and instead that a simple technique called scale-then-privatize (which does not achieve unbiased second moments) has more desirable theoretical behaviors and outperforms all other variants we study on a small-scale language model training task. We additionally argue that scale-then-privatize causes the noise addition to better match the application of correlated noise mechanisms which are more desirable to use in practice.
\end{abstract}

\section{Introduction}

Language models often have sparse gradients in training, and it is now well-understood that using adaptive optimizers leads to better training performance for these models \cite{zhao2025deconstructingmakesgoodoptimizer}. At a high level, adaptive optimizers maintain $\bnu$, an estimate of the average coordinate-wise square of the gradients, and rather than use a constant learning rate, use a per-coordinate learning proportional to $1/\sqrt{\bnu}$ (where $\sqrt{\bnu}$ denotes the per-coordinate square root; the application of per-coordinate learning rate is also commonly called preconditioning). Intuitively, $\bnu$ represents the information learned about the geometry of the gradients seen so far, and the use of per-coordinate learning rates is an attempt to adapt the gradients to this geometry.

When training a model with differential privacy (DP) \cite{DMNS} and a non-adaptive optimizer, usually noise is added to the gradients and then the noisy gradients are passed to the non-adaptive optimizer in a black-box manner. One can analogously train with DP using an adaptive optimizer in a black-box manner, but doing so indirectly noises $\bnu$. Roughly speaking, this noise causes the coordinates of $\bnu$ to be more similar in magnitude, i.e. impedes the ability of the adaptive optimizer to learn the geometry of the gradients and makes it behave more similarly to a non-adaptive optimizer.  Consequently, even for tasks where non-private adaptive optimizers outperform non-adaptive optimizers, for DP adaptive and non-adaptive optimizers may achieve similar performance \cite{tang2023dpadambc}. In turn, many recent works (e.g., \cite{asi21privateadaptive, li22side, li2023differentially, tang2023dpadambc, kalinin2025continual}) have proposed new variants of private adaptive optimizers that aim to recover the benefits of adaptive optimizers by restoring the ability of the adaptive optimizer to learn the geometry in the presence of noise. 

In spite of the many proposed variants, our understanding of these variants remains limited in a few aspects. First, the empirical evaluation in many of these papers is restricted to tasks such as CIFAR or text classification tasks such as SNLI/QNLI, which are much simpler than the token prediction tasks language models are trained to solve in practice. Hence it is not clear if the conclusions of existing empirical studies extend to production-scale fine-tuning of language models with DP. Second, with the exception of \cite{kalinin2025continual}, most work only focuses on adding independent noise in different iterations, but it is often preferable in practice to use noise that is instead correlated across iterations \cite{mcmahan2024hassle}. Even \cite{kalinin2025continual} focuses primarily on the role of correlated noise in the estimation of $\bnu$, and less on the interplay between correlated noise and changing per-coordinate learning rates in \adam.

\subsection{Our Contributions}

We survey several private variants of adaptive optimizers, detailed in \cref{sec:variants}. We analyze these variants both by studying their theoretical underpinnings, and by conducting experiments comparing their effectiveness for training a Transformer language model (TinyBERT) on a token prediction task.

\mypar{Theoretical contributions (\cref{sec:theory})} We build theoretical intuition about the interplay between noise addition and adaptive optimizers. We propose several insights on how to design private adaptive optimizers and give quantitative justification for each. Our main insights are:

\begin{itemize}
    \item \mypar{Unbiased $\bnu$ is not necessary} 
    Combining DP gradients and adaptive optimizers in a black-box fashion gives biased estimates of $\bnu$, and previous works have suggested improving private adaptive optimizers by using an unbiased estimate instead. We argue that this is somewhat misguided, as these approaches do not distinguish variance in the data from variance due to DP noise. We further argue that a previously proposed technique which we call \textit{scale-then-privatize}, which uses $\bnu$ to clip and noise the gradients in a non-isotropic manner, is better because it both learns the non-private geometry and causes the noisy gradient distribution to match this geometry.
    \item \mypar{Two regimes for unbiased $\bnu$} We further argue that using $\bnu$ which is an unbiased estimate of $\bnu^*$, the value of $\bnu$ we'd arrive at with the same gradients and no noise, can be harmful in certain parameter regimes. When using unbiased but noisy estimates of $\bnu$, we show that if the dimension is sufficiently large (as a function of the noise level and batch size), a constant fraction of the coordinates of $\bnu$ (which is estimating a weighted average of squared values, i.e. non-negative values), will be negative. While this can be corrected by e.g. taking only the positive part of $\bnu$, even with these corrections this can lead to undesirable learning behaviors. This observation can explain a disparity between empirical results in the literature operating with lower-dimensional tasks and our empirical results on transformer training.
    \item \mypar{Scale-then-privatize benefits correlated noise} We lastly consider the interplay between correlated noise in private optimizers and adaptive optimizers. Correlated noise, also known as \dpmf, attempts to cancel out the noise added in one iteration in subsequent iterations, to reduce the total amount of noise added across the learning process. However, when using an adaptive optimizer the per-coordinate learning rate changes across iterations, which may impede this cancellation. We show that scale-then-privatize causes the effective noise added to each \textit{update} (rather than each gradient, i.e. the noise after rescaling by the per-coordinate learning rates) to be roughly isotropic, suggesting that noise cancellation should be about as effective as in the non-adaptive setting, regardless of changes in the per-coordinate learning rate over time.
\end{itemize}

\mypar{Empirical results (\cref{sec:empirical})} We conduct experiments where we train models using the different variants consider. We first observe that in a one-dimensional setting, a technique called independent moment estimation significantly outperforms the baseline of using adaptive optimizers in a black-box manner. However, when we move to a token prediction task where we train TinyBERT, a transformer model with only a few million trainable parameters, most variants we consider do worse than the black-box approach. 
This supports our theoretical intuition that approaches targeting unbiased estimates of $\bnu$ can break down once the dimension is sufficiently high, and furthermore that this occurs at model sizes much smaller than those used in production settings. We also demonstrate that unlike the other variants, scale-then-privatize outperforms the black-box approach for this task. In conjunction with our theoretical intuition, our empirical results suggest that the empirical results in past works may not generalize even to small-scale token prediction tasks that better resemble model training in production settings, and the principles used to design private adaptive optimizers in past work may need to be reconsidered.

\section{Background}\label{sec:background}

\paragraph{Notation} We first establish some notation, which we will use throughout to describe learning algorithms. We let $\bfg^2 = \bfg \odot \bfg$ where $\odot$ denotes the element-wise product of two vectors, and $\bfu / \bfv$ denote the element-wise division of two vectors. For a vector $\bfa$ and a scalar $b$, we use $\bfa + b$ as shorthand for $\bfa + b \cdot \boldone$, and $\max\{\bfa, b\}$ denotes the element-wise max of $\bfa$ and $b$. We let $[T] = (1, \dots, T)$, and let $\bfg_{t, j} \in \R^\mdim$ for $t \in [T]$ and $j \in [B]$ be the $j$th gradient from the minibatch of size $B$ seen on round $t$. Then, $G_t = \{\bfg_{1,j} \mid j \in [B]\}$ represents the set of $B$ gradients for round $t$. We define the clipping operator
\[
\clip{\bfv} \coloneqq \bfv \cdot \min\{1, \clipnorm / \ltwo{\bfv}\}
\]
which clips gradients to have norm at most $\clipnorm$. Some symbols representing algorithm internal state or temporary values may have different meanings in different algorithms, for example we write
\[
\bfg_t = \frac{1}{B} \sum_{\bfg_{t,j} \in G_t} \bfg_{t,j}
\qquad \text{or} \qquad
\bfg_t = \frac{1}{B} \sum_{\bfg_{t,j} \in G_t}  \clip{\bfg_{t, j}}
\]
for non-private vs private algorithms respectively.  

\subsection{Private training via \dpsgd and \dpmf}

An algorithm $\calA$ is $(\epsilon, \delta$)-DP if given two datasets $D, D'$ that are adjacent according to some adjacency definition, and any set of possible outputs $S$ we have:

\[\Pr[A(D) \in S] \leq e^\epsilon \Pr[A(D') \in S] + \delta.\]

We focus on the zero-out adjacency as defined in \cite{ponomareva23dpfy}: $D, D'$ are adjacent if $D'$ is equal to $D$ (or vice-versa) except one example is replaced with a placeholder $\bot$, whose gradient is zero everywhere.\footnote{For production uses where one user may contribute multiple examples, our results extend naturally to user-level DP, see for example \cite{charles2024finetuninglargelanguagemodels}.} A standard DP for answering a vector-valued query on a dataset is the Gaussian mechanism: If $f$ is a query such that for any adjacent $D, D'$ we have $\ltwo{f(D) - f(D')} \leq \Delta$, the Gaussian mechanism with noise multiplier $\sigma$ answers $f$ by returning $f(D) + \bfz$, where $\bfz \sim N(\boldzero, \sigma^2 \Delta^2 \mathbb{I})$. 

\dpsgd \cite{song2013stochastic, BST14, DP-DL} is a canonical private learning algorithm which satisfies $(\epsilon, \delta)$-DP. In \dpsgd, given in \cref{fig:dpsgd}, in each round $t$ we sample a batch of examples from the dataset $D$ and compute their gradients $G_t$ on the current model, clip the gradients to have bounded $\ell_2$-norm, and add independent Gaussian noise to their average. We then use this as a noisy gradient in \sgd, or another first-order method such as \adam, as the standard privacy guarantees of \dpsgd do not depend on the first-order method by the post-processing property of DP.

\begin{figure}
\begin{algorithm}[H]
\caption{The \dpsgd/\dpmf algorithms}
\label{fig:dpsgd}
\textbf{Inputs:} Stream of batch gradients $(G_1 = \{\bfg_{1,j}\}, G_2 = \{\bfg_{2,j}\}, \ldots, G_T \{\bfg_{T,j}\})$, number of rounds $T$, batch size $B$, clip norm $\clipnorm$, noise multiplier $\sigma$, initial model $\bftheta_0$, learning rate $\eta$, noise-correlating matrix $\Cinv$ (for \dpsgd, $\Cinv = \bfI$).
\begin{algorithmic}[1]
\State $\clip{\bfv} \coloneqq \bfv \cdot \min\{1, \clipnorm / \ltwo{\bfv}\}$
\For{$t \in [T]$}
\State $\bfg_t = \frac{1}{B} \sum_{\bfg_{t,j} \in G_t}  \clip{\bfg_{t, j}}$
\State $\bfz_t \sim N(\boldzero, \bfI) \in \R^\mdim$
\State $\tilde{\bfg}_t = \bfg_t + \frac{\clipnorm\sigma}{B} (\Cinv \bfz)_t$ \Comment{$\bfz = (\bfz_1, \bfz_2, \ldots) \in \R^{T \times \mdim}$}
\State $\bftheta_t = \bftheta_{t-1} - \eta \tilde{\bfg}_t$ (or another first-order update)
\EndFor
\end{algorithmic}
\end{algorithm}
\vspace{-1.2em}
\end{figure}

\dpmf \cite{kairouz2021practical}, or equivalently, \dpsgd with correlated noise, is also given in \cref{fig:dpsgd}. In this variant of \dpsgd the noise used in different rounds is correlated, so that some of the noise added in one round is cancelled out in subsequent rounds. The amount of cancellation is specified by a lower triangular noise correlating matrix\footnote{We use $\Cinv$ for consistency with other works; the matrix $\bfC$ is called the strategy matrix, and plays an important role in the privacy analysis of \dpmf, but will not be critical to our work.} $\Cinv\in \R^{T \times T}$,  which produces noise $(\Cinv \bfz)_t = \sum_{j=1}^t \Cinv_{t, j} \bfz_j$ in round $t$. The case $\Cinv = \bfI$ retrieves the independent noise of \dpsgd. Typically, $\Cinv$ is chosen via a suitable matrix factorization, e.g. \cite{denisov2022improved, choquette2022multi}. Typically the main diagonal of $\Cinv$ will be positive (often 1), with negative values below the main diagonal. For example,
\[
\Cinv = \begin{bmatrix}
1   & 0   & 0 \\
-0.5 & 1   & 0 \\
0   & -0.5 & 1 
\end{bmatrix}
\]
adds one unit of noise on each round, and then subtracts off half of this noise on the subsequent iteration (while also adding one unit of new IID noise).

The privacy guarantees of the Gaussian mechanism as well as the privacy analysis of \dpsgd and \dpmf via privacy analysis of the Gaussian mechanism are by now well-understood. For the most part precise privacy guarantees and parameters are immaterial to the discussions in this paper, and we will focus on analyzing \dpadam with a fixed noise multiplier agnostic of the corresponding privacy parameters.

\subsection{Background on \adagrad/\adam}

While there are a number of adaptive optimizers of interest, for simplicity we will primarily focus on \adam (of \cite{kingma2014adam}), and occasionally also consider \adagrad (of \cite{mcmahan2010adaptive,duchi2011adaptive, mcmahan10boundopt}); many of our theoretical observations and experimental conclusions should readily transfer to other adaptive optimizers. Pseudocode for both algorithms is given in \cref{fig:adam} and \cref{fig:adagrad}.

\begin{figure}[H]
\begin{minipage}[t]{0.48\textwidth}

\begin{figure}[H]
\begin{algorithm}[H]
\caption{The \adam algorithm}
\label{fig:adam}
\textbf{Inputs:} Stream of batch gradients $(G_1 = \{\bfg_{1,j}\}, G_2 = \{\bfg_{2,j}\}, \ldots, G_T = \{\bfg_{T,j}\})$, 
initial model $\bftheta_0 \in \R^\mdim$, base learning rate $\eta$, momentum parameters $\beta_1, \beta_2$, stability constant $\stability$.
\begin{algorithmic}[1]
\State $\bmu_0, \bnu_0 = \boldzero$
\For{$t \in [T]$}
\State $\bfg_t = \frac{1}{B} \sum_{\bfg_{t,j} \in G_t} \bfg_{t,j}$
\State $\bmu_t = \beta_1 \bmu_{t-1} + (1 - \beta_1) \bfg_t$
\State $\hat{\bmu}_t = \bmu_t/(1 - \beta_1^t)$
\State $\bnu_t = \beta_2 \bnu_{t-1} + (1 - \beta_2) \bfg_t^2$
\State $\hat{\bnu}_t = \bnu_t/(1 - \beta_2^t)$
\State $\bftheta_t = \bftheta_{t-1} - \eta \cdot \hat{\bmu}_t / (\sqrt{\hat{\bnu}_t} + \stability)$
\EndFor

\end{algorithmic}
\end{algorithm}
\end{figure}
\end{minipage}
\hfill
\begin{minipage}[t]{0.48\textwidth}
\begin{figure}[H]
\begin{algorithm}[H]
\caption{The \adagrad algorithm}
\label{fig:adagrad}
\textbf{Inputs:} Stream of batch gradients $(G_1 = \{\bfg_{1,j}\}, G_2 = \{\bfg_{2,j}\}, \ldots, G_T \{\bfg_{T,j}\})$, initial model $\bftheta_0$, base learning rate $\eta$.
\begin{algorithmic}[1]
\State $\bnu_0 = \boldzero$
\For{$t \in [T]$}
\State $\bfg_t = \frac{1}{B} \sum_{\bfg_{t,j} \in G_t} \bfg_{t,j}$
\State $\bnu_t = \bnu_{t-1} + \bfg_t^2$
\State $\bftheta_t = \bftheta_{t-1} - \eta \cdot \bfg_t / \sqrt{\bnu_t}$.
\EndFor

\end{algorithmic}
\end{algorithm}
\end{figure}
\end{minipage}
\end{figure}

The key idea of both is to maintain $\bnu$, a (weighted) sum of all squared gradients $\bfg^2$ seen so far, and use (roughly) $1/\sqrt{\bnu}$ as a vector of per-coordinate learning rates. While ideally for $\mdim$-dimensional gradients we would maintain a $\mdim \times \mdim$ matrix estimating $\mathbb{E}_\bfg[\bfg\bfg^\top]$, due to memory limitations in practice it is more common to maintain an estimate of $\mathbb{E}_\bfg[\bfg^2]$. We hence give the pseudocode for the latter throughout. 

\adagrad chooses $\bnu$ to be exactly the sum of squared gradients. Intuitively, this retrieves the standard theoretical choice of learning rate decay $1/\sqrt{t}$ for \sgd in the one-dimensional setting if the gradient is always a constant, and then maintains this behavior even if we insert any number of zero gradients to the input stream, or scale all the gradients by a constant. \adam uses a sum of decaying squared gradients, and normalizes the sum to get $\hat{\bnu}$ before computing the per-coordinate learning rates using $\hat{\bnu}$ instead of $\bnu$. Intuitively, \adam recovers \adagrad's behavior of using larger learning rate in coordinates that have been updated less, but (i) the decay enables it to adapt to changes in the magnitudes throughout the optimization process, whereas \adagrad may be stuck using smaller learning rates in coordinates that have previously seen many large gradients, and (ii) the normalization means it retrieves behavior closer to constant learning rates rather than decaying learning rates, which may be preferable in practice.

\subsubsection{Square-Then-Average} As written, \adagrad and \adam first average the gradients in a batch before squaring. One could alternatively square the individual gradients before averaging. If we view the choice of $1/\sqrt{\bnu}$ as trying to retrieve the $1/\sqrt{M}$ learning rate that is standard for $M$-smooth losses, one could argue that the variance of gradients is a proxy for this squared smoothness, and the average squared gradient is a better estimate for the variance than the squared average gradient. While exploring the benefits of ``square-then-average'' rather than ``average-then-square'' is an interesting direction, we focus on average-then-square for two reasons. First, average-then-square is the variant used by the original \adam paper \cite{kingma2014adam} and the one commonly used in practice, so by focusing on this variant our results are more directly applicable to \adam's use in practice. Second, the decision to use square-then-average rather than average-then-square is relevant even for non-private learning, and hence somewhat outside the scope of this paper.

\section{Survey of \dpadam Variants}\label{sec:variants}

Here we define and discuss some proposed DP variants of \adam. It is usually straightforward to define the analogous versions of these algorithms for other adaptive optimizers such as \adagrad. For simplicity we present the algorithms using independent noise (\dpsgd), but they can be analogously defined using \dpmf in all cases by letting $\bfz_t$ be the $t$th value of correlated noise, i.e. using $(\Cinv \bfz)_t$ instead of $\bfz_t$ as in \cref{fig:dpsgd}. 

We first give pseudocode in \cref{fig:pseudocode} for all the variants and then discuss them in more detail in the following subsections. \cref{fig:adam-dup} restates the non-private \adam algorithm for convenience, \cref{fig:adam-pp} gives \dpadam via post-processing with the differences  from \adam highlighted in blue, and then in the remaining private variants of \adam we highlight the differences with \dpadam via post-processing rather than non-private \adam.

\begin{figure}[H]
\begin{minipage}[t]{0.48\textwidth}

\begin{figure}[H]
\begin{algorithm}[H]
\caption{The Adam algorithm}
\label{fig:adam-dup}
\begin{algorithmic}[1]
\State $\bmu_0, \bnu_0 = \boldzero$
\For{$t \in [T]$}
\State $\bfg_t = \frac{1}{B} \sum_{\bfg_{t,j} \in G_t} \bfg_{t,j}$
\State $\bmu_t = \beta_1 \bmu_{t-1} + (1 - \beta_1) \bfg_t$
\State $\hat{\bmu}_t = \bmu_t/(1 - \beta_1^t)$
\State $\bnu_t = \beta_2 \bnu_{t-1} + (1 - \beta_2) \bfg_t^2$
\State $\hat{\bnu}_t = \bnu_t/(1 - \beta_2^t)$
\State $\bftheta_t = \bftheta_{t-1} - \eta \cdot \hat{\bmu}_t / (\sqrt{\hat{\bnu}_t} + \epsilon_s)$
\EndFor

\end{algorithmic}
\end{algorithm}
\end{figure}
\end{minipage}
\hfill
\begin{minipage}[t]{0.48\textwidth}
\begin{figure}[H]
\begin{algorithm}[H]
\caption{Post-processing (\cref{sec:adam-pp})}
\label{fig:adam-pp}
\begin{algorithmic}[1]
\State $\bmu_0, \bnu_0 = \boldzero$
\For{$t \in [T]$}
\State \bluetext{$\bfg_t = \frac{1}{B} \sum_{\bfg_{t,j} \in G_t} \clip{\bfg_{t,j}}$}
\State \bluetext{$\bfz_t \sim N(\boldzero, \sigma^2 \mathbb{I}_p)$}
\State \bluetext{$\tilde{\bfg}_t = \bfg_t + \frac{\clipnorm}{B} \bfz_t$}
\State $\bmu_t = \beta_1 \bmu_{t-1} + (1 - \beta_1) \bluetext{\tilde{\bfg}_t}$
\State $\hat{\bmu}_t = \bmu_t/(1 - \beta_1^t)$
\State $\bnu_t = \beta_2 \bnu_{t-1} + (1 - \beta_2) \bluetext{\tilde{\bfg}_t^2}$
\State $\hat{\bnu}_t = \bnu_t/(1 - \beta_2^t)$
\State $\bftheta_t = \bftheta_{t-1} - \eta \cdot \hat{\bmu}_t / (\sqrt{\hat{\bnu}_t} + \epsilon_s)$
\EndFor
\end{algorithmic}
\end{algorithm}
\end{figure}
\end{minipage}
\end{figure}

\vspace{-0.5in}

\begin{figure}[H]
\begin{minipage}[t]{0.48\textwidth}

\begin{figure}[H]
\begin{algorithm}[H]
\caption{Independent moment estimation (\cref{sec:adam-ime})}
\label{fig:adam-ime}
\begin{algorithmic}[1]
\State $\bmu_0, \bnu_0 = \boldzero$
\For{$t \in [T]$}
\State $\bfg_t = \frac{1}{B} \sum_{\bfg_{t,j} \in G_t} \clip{\bfg_{t,j}}$
\State $\bfz_{t,1}, \bfz_{t,2} \sim N(\boldzero, \sigma^2 \mathbb{I}_{p})$
\State $\bmu_t = \beta_1 \bmu_{t-1} + (1 - \beta_1) \left(\bfg_t + \frac{\bluetext{\sqrt{2}}\clipnorm}{B}\bfz_{t,1}\right)$
\State $\hat{\bmu}_t = \bmu_t / (1 - \beta_1^t)$
\State $\bnu_t = \beta_2 \bnu_{t-1} + (1 - \beta_2) \bluetext{(\bfg_t^2 + \frac{\sqrt{2}(2B+1)\clipnorm^2}{B^2} \bfz_{t,2})}$
\State $\hat{\bnu}_t = \bnu_t / (1 - \beta_2^t)$
\State $\bftheta_t = \bftheta_{t-1} - \eta \cdot \hat{\bmu}_t / (\sqrt{\max\{\hat{\bnu}_t, \boldzero\}} + \epsilon_s)$
\EndFor
\end{algorithmic}
\end{algorithm}
\end{figure}
\end{minipage}
\hfill
\begin{minipage}[t]{0.48\textwidth}
\begin{figure}[H]
\begin{algorithm}[H]
\caption{Bias correction (\cref{sec:adambc})}
\label{fig:adambc}
\begin{algorithmic}[1]
\State $\bmu_0, \bnu_0 = \boldzero$
\For{$t \in [T]$}
\State $\bfg_t = \frac{1}{B} \sum_{\bfg_{t,j} \in G_t} \clip{\bfg_{t,j}}$
\State $\bfz_t \sim N(\boldzero, \sigma^2 \mathbb{I}_p)$
\State $\tilde{\bfg}_t = \bfg_t + \frac{\clipnorm}{B} \bfz_t$
\State $\bmu_t = \beta_1 \bmu_{t-1} + (1 - \beta_1) \tilde{\bfg}_t$
\State $\hat{\bmu}_t = \bmu_t/(1 - \beta_1^t)$
\State $\bnu_t = \beta_2 \bnu_{t-1} + (1 - \beta_2) \tilde{\bfg}_t^2$
\State $\hat{\bnu}_t  = \bnu_t/(1 - \beta_2^t)$
\State $\bftheta_t = \bftheta_{t-1} - \eta \cdot \hat{\bmu}_t / \sqrt{\max\{\hat{\bnu}_t \bluetext{- \frac{\clipnorm^2 \sigma^2}{B^2}}, \epsilon_s^2\}}$
\EndFor
\end{algorithmic}
\end{algorithm}
\end{figure}
\end{minipage}
\end{figure}

\vspace{-0.5in}

\begin{figure}[H]
\begin{minipage}[t]{0.48\textwidth}

\begin{figure}[H]
\begin{algorithm}[H]
\caption{Scale-then-privatize (\cref{sec:adam-pc})}
\label{fig:adam-pc}
\begin{algorithmic}[1]
\State $\bmu_0, \bnu_0, \bluetext{\hat{\bnu}_0} = \boldzero$
\For{$t \in [T]$}
\State \bluetext{$\bfs_t = 1 / (\sqrt{\hat{\bnu}_{t-1}} + \epsilon_{s_1})$}
\State $\bfg_t = \frac{1}{B} \sum_{\bfg_{t,j} \in G_t} \bluetext{%
\clip{\bfs_t \odot \bfg_{t,j}}}$\label{line:scale-then-clip}
\State $\bfz_t \sim N(\boldzero, \sigma^2 \mathbb{I}_p)$
\State $\tilde{\bfg}_t = (\bfg_t + \frac{\clipnorm}{B} \bfz_t) \bluetext{/ \bfs_t}$\label{line:scaled-noise}
\State $\bmu_t = \beta_1 \bmu_{t-1} + (1 - \beta_1) \tilde{\bfg}_t$
\State $\hat{\bmu}_t = \bmu_t/(1 - \beta_1^t)$
\State $\bnu_t = \beta_2 \bnu_{t-1} + (1 - \beta_2) \tilde{\bfg}_t^2$
\State $\hat{\bnu}_t = \bnu_t/(1 - \beta_2^t)$
\State $\bftheta_t = \bftheta_{t-1} - \eta \cdot \hat{\bmu}_t / (\sqrt{\hat{\bnu}_t} + \epsilon_{s_2})$
\EndFor
\end{algorithmic}
\end{algorithm}
\end{figure}
\end{minipage}
\hfill
\begin{minipage}[t]{0.48\textwidth}
\begin{figure}[H]
\begin{algorithm}[H]
\caption{Side Information (\cref{sec:adam-side})}
\label{fig:adam-side}
\begin{algorithmic}[1]
\State $\bmu_0 = \boldzero$
\For{$t \in [T]$}
\State $\bfg_t = \frac{1}{B} \sum_{\bfg_{t,j} \in G_t} \clip{\bfg_{t,j}}$
\State $\bfz_t \sim N(\boldzero, \sigma^2 \mathbb{I}_p)$
\State $\tilde{\bfg}_t = \bfg_t + \frac{\clipnorm}{B} \bfz_t$
\State $\bmu_t = \beta_1 \bmu_{t-1} + (1 - \beta_1) \tilde{\bfg}_t$
\State $\hat{\bmu}_t = \bmu_t/(1 - \beta_1^t)$
\State \bluetext{Compute $\hat{\bnu}_t$ using side information}
\State $\bftheta_t = \bftheta_{t-1} - \eta \cdot \hat{\bmu}_t / \sqrt{\hat{\bnu}_t}$
\EndFor
\end{algorithmic}
\end{algorithm}
\end{figure}
\end{minipage}
\caption{Pseudocode for \adam and its private variants.}\label{fig:pseudocode}
\end{figure}

\subsection{Post-Processing (``Standard'' \dpadam)}\label{sec:adam-pp}

Arguably the simplest and most common way to make \adam private is post-processing (\cref{fig:adam-pp}): privatize the computation of $\bfg_t$, and then use the privatized $\bfg_t$ in \adam in a black-box manner.

The main issue with the algorithm is that while the Gaussian noise is mean-zero and hence $\bmu_t$ is unbiased, $\bnu_t$ is biased. In particular, if $\bnu_t^*$ is the value of $\bnu_t$ one would get from the unnoised \cref{fig:adam} for the same gradient stream\footnote{Note that this is \textit{not} the same as the value of $\bnu_t$ one would arrive at with non-private \adam, because non-private \adam would take a different trajectory and hence receive a different gradient stream.}, then we can compute the bias for each term as follows. For $\bnu_1$ we have:

\begin{align*}
\mathbb{E}[\bnu_1] = (1 - \beta_2)\mathbb{E}[\tilde{\bfg}_1^2] &= (1-\beta_2)(\bfg_t^2 + \frac{\clipnorm^2}{B^2}\mathbb{E}[\bfz_t^2])\\
&= \bnu_t^* + (1-\beta_2)\frac{\clipnorm^2 \sigma^2}{B^2}. 
\end{align*}

Now by induction assume $\mathbb{E}[\bnu_{t-1}] = \bnu_{t-1}^* + (1-\beta_2^{t-1}) \frac{\clipnorm^2 \sigma^2}{B^2}$. Then:

\begin{align*}
\mathbb{E}[\bnu_t] &= \beta_2 \mathbb{E}[\bnu_{t-1}] + (1 - \beta_2)\mathbb{E}[\tilde{\bfg}_t^2] \\
&= \beta_2 \bnu_{t-1}^* + \beta_2(1-\beta_2^{t-1}) \frac{\clipnorm^2 \sigma^2}{B^2} + (1-\beta_2) (\bfg_t^2 + \frac{\clipnorm^2 \sigma^2}{B^2})\\
&= \beta_2 \bnu_{t-1}^*+(1-\beta_2)\bfg_t^2 + (1-\beta_2^t)\frac{\clipnorm^2 \sigma^2}{B^2}.\\
&= \bnu_t^* + (1-\beta_2^t)\frac{\clipnorm^2 \sigma^2}{B^2}.
\end{align*}

Hence by induction, each $\bnu_t$ has bias $(1-\beta_2^t)\frac{\clipnorm^2 \sigma^2}{B^2}\cdot \boldone$, i.e. $\hat{\bnu}_t$ has bias $\frac{\clipnorm^2 \sigma^2}{B^2}\cdot \boldone$. For \dpmf a slightly more nuanced calculation is involved, which we give in \cref{sec:dpmf-bc}.

The effect of this bias effectively causes all coordinates where $\bnu_t^*$ is sufficiently small, i.e. where gradients have small magnitude or have non-zero mass infrequently, to use a much smaller per-coordinate learning rate $1/\sqrt{\bnu_t}$. \citet{tang2023dpadambc} argue that this makes \adam resemble \dpsgd with momentum, since \adam with $\bnu_t \propto \boldone$ retrieves \sgd with momentum and the noise causes $\bnu_t$'s coordinates to be more uniform.

\subsection{Independent Moment Estimation}\label{sec:adam-ime}

One way to fix the bias in $\bnu_t$ is independent moment estimation (\cref{fig:adam-ime}), as defined in \cite{kalinin2025continual}: we separately noise $\bmu_t$ and $\bnu_t$. Equivalently, we treat $(\bfg_t, c \bfg_t^2)$ for $c \in \R^+$ as a $2\mdim$-dimensional vector to be noised in each round. Independent moment estimation adds i.i.d Gaussian noise to this vector, and then uses the first $\mdim$ coordinates to update $\bmu_t$ in \adam and the second $\mdim$ coordinates to update $\bnu_t$. Here, $c$ is a scaling term that trades off which of $\bmu_t, \bnu_t$ we want to add more noise to: When $c$ is high the $\ell_2$-sensitivity of this vector is proportional to $c$, so $\bmu_t$ is over-noised. When $c$ is low the sensitivity is $\approx \clipnorm$, but $\bfg_t^2$ is scaled by $c$ so it is over-noised.

For simplicity we consider independent moment estimation with $c = 1$. Under the zero-out adjacency, the sensitivity of $\bfg_t$ remains $\clipnorm/B$, and the sensitivity of $\bfg_t^2$ is bounded as:
\begin{align*}
&\max_{\bfg_{t,1}, \bfg_{t,2}, \ldots, \bfg_{t,B} : 
\forall j \ltwo{\bfg_{t,j}} \leq \clipnorm} 
 \ltwo{\left(\frac{1}{B} \sum_{j=1}^B \bfg_{t,j}\right)^2 - \left(\frac{1}{B} \sum_{j=1}^{B-1} \bfg_{t,j}\right)^2}\\
= &\max_{\bfg_{t,1}, \bfg_{t,2}, \ldots, \bfg_{t,B} : \forall j \ltwo{\bfg_{t,j}} \leq \clipnorm} \ltwo{\frac{2}{B^2} \left(\sum_{t=1}^{B-1} \bfg_{t,j}\right) \odot \bfg_{t,B} + \frac{1}{B^2} \bfg_{t,B}^2}\\
\leq &\max_{\bfg_{t,1}, \bfg_{t,2}, \ldots, \bfg_{t,B} : \forall j \ltwo{\bfg_{t,j}} \leq \clipnorm} \frac{2}{B^2}\ltwo{ \left(\sum_{t=1}^{B-1} \bfg_{t,j}\right) \odot \bfg_{t,B}} + \frac{1}{B^2}\ltwo{ \bfg_{t,B}^2}\\
= &\max_{\bfg_{t,1}, \bfg_{t,2}, \ldots, \bfg_{t,B} : \forall j \ltwo{\bfg_{t,j}} \leq \clipnorm} \frac{2}{B^2}\ltwo{ \sum_{t=1}^{B-1} \bfg_{t,j} \odot \bfg_{t,B}} + \frac{1}{B^2}\ltwo{ \bfg_{t,B}^2}\\
\leq &\max_{\bfg_{t,1}, \bfg_{t,B} :  \ltwo{\bfg_{t,1}}, \ltwo{\bfg_{t,B}} \leq \clipnorm} \frac{2(B-1)}{B^2}\ltwo{\bfg_{t,1} \odot \bfg_{t,B}} + \frac{1}{B^2}\ltwo{ \bfg_{t,B}^2}\\
= &\max_{\bfg_{t,1}, \bfg_{t,2}, \ldots, \bfg_{t,B} : \forall j \ltwo{\bfg_{t,j}} \leq \clipnorm} \frac{2(B-1)\clipnorm^2}{B^2} + \frac{\clipnorm^2}{B^2} = \frac{(2B-1)\clipnorm^2}{B^2},
\end{align*}
The last inequality uses the following inequality:

\begin{align*}
\ltwo{\bfg_{t,1} \odot \bfg_{t,B}} &= \sqrt{\sum_{i \in [\mdim]} \bfg_{t,1}(i)^2 \bfg_{t,B}(i)^2} \\
&\leq \sqrt{\left(\sum_{i \in [\mdim]} \bfg_{t,1}(i)^2\right) \left(\sum_{i \in [\mdim]} \bfg_{t,B}(i)^2\right)}\\
&= \sqrt{\sum_{i \in [\mdim]} \bfg_{t,1}(i)^2} \sqrt{\sum_{i \in [\mdim]} \bfg_{t,B}(i)^2}\\
&= \ltwo{\bfg_{t,1}} \ltwo{\bfg_{t,B}}\\
&\le \clipnorm^2.
\end{align*}

This sensitivity bound is tight in the case where all $\bfg_{t,j}$ are the same. By composition (which states that the combination of two Gaussian mechanisms with noise multipliers $\sqrt{2}\sigma$ satisfiy the privacy guarantees of one Gaussian mechanism with noise multiplier $\sigma$), to achieve the same privacy as e.g. \dpadam via post-processing, it then suffices to add noise with standard deviation $\frac{\sqrt{2}\clipnorm}{B}$ to $\bfg_t$ and standard deviation $\frac{\sqrt{2}(2B+1)\clipnorm^2}{B^2}$ to $\bfg_t^2$ and this is tight again in the case where all $\bfg_{t,j}$ are the same. Hence we arrive at the variant of \adam defined in \cref{fig:adam-ime}. 

\citet{kalinin2025continual} observed that under the \textit{replace adjacency}\footnote{In the replace adjacency $D$ and $D'$ are adjacent if we can replace an example in $D$ with any other example (not necessarily the special example $\bot$ used in the zero-out adjacency) to get $D'$.}, the sensitivity of the vector $(\bfg_t, c \bfg_t^2)$ could be the same as the sensitivity of the vector $\bfg_t$ for $c$ as large as $1/2 \clipnorm^2$. They term noising this vector (with a reduced sensitivity bound) \textit{joint} moment estimation. The intuitive reason there is no increase in sensitivity is when bounding the sensitivity of $\bfg_t$, the ``worst-case'' pair of values $\bfg_t, \bfg_t'$ a gradient takes on for two adjacent datasets satisfies $\bfg_t = -\bfg_t'$. But in this case, $\bfg_t^2 = \bfg_t'^2$, i.e. concatenating $c \bfg_t^2$ doesn't increase the sensitivity. Furthermore, if $c \leq 1/2 \clipnorm^2$, \cite{kalinin2025continual} shows that $\bfg_t = -\bfg_t'$ remains the worst-case for the sensitivity of the pair $(\bfg_t, c \bfg_t^2)$\footnote{This bound on $c$ is for the case where $\mdim > 1$ and we use \dpsgd. If the dimension is $\mdim = 1$ they show an improved bound. Furthermore, if using \dpmf instead of \dpsgd for noise addition, the bound also depends on the choice of factorization.}. However, for the zero-out adjacency we focus on the worst case is $\bfg_t = \boldzero$ instead of $\bfg_t = -\bfg_t'$, i.e. in the zero-out adjacency strictly more noise is required if additionally noising $\bfg_t^2$. 

Unlike post-processing, the noisy second moment estimate $\bnu_t$ is not necessarily positive because the noise $\bfz_{t,2}$ added to it is not squared. Hence we need to take the positive part of $\bnu_t$ before taking the square root. There are other ways to correct this to avoid taking the square root of a negative number, for example by using $\sqrt{\max\{\bnu_t, \stability^2\}}$ instead of $\sqrt{\max\{\bnu_t, \boldzero\}} + \stability \cdot \bold 1$. We do not believe the behaviors of these variants substantially differ, as they are never more than a constant factor apart from each other.

\subsection{Bias Correction}\label{sec:adambc}

Bias correction (\cref{fig:adambc}), proposed by \citet{tang2023dpadambc}, handles the bias in $\hat{\bnu}_t$ by directly subtracting the bias $\frac{\clipnorm^2 \sigma^2}{B^2}$ from it.

As with independent moment estimation, bias correction gives an estimate of the second moment $\hat{\bnu}_t - \frac{\clipnorm^2 \sigma^2}{B^2}$ that is unbiased, but can be negative, and hence we need to take the positive part of it before taking its square root to compute the per-coordinate learning rates.

\subsection{Scale-Then-Privatize}\label{sec:adam-pc}

All approaches we have discussed so far clip gradients in the standard $\ell_2$ norm. However, $\hat{\bnu}_t$ defines a different norm/geometry, such that \adam taking a gradient step using different per-coordinate learning rates $\eta/\sqrt{\hat{\bnu}_t}$ in the ``standard'' geometry is equivalent to taking a gradient step using constant learning rate $\eta$ in this geometry. We could instead consider clipping and noising gradients in this geometry, which we refer to as \textit{scale-then-privatize} (\cref{fig:adam-pc}). Ideally we would use $\hat{\bnu}_t$, which contains the most ``up-to-date'' estimate of the geometry, to scale each $\bfg_{t,j}$, but this of course introduces a circular dependence between the parameters. 
We instead use $\hat{\bnu}_{t-1}$ to scale $\bfg_{t,j}$. For small values of $t$,  $\hat{\bnu}_t$ and $\hat{\bnu}_{t-1}$ may differ largely, but as $t$ increases $\hat{\bnu}_{t-1}$ will become a reasonably good estimate of $\hat{\bnu}_t$.

Here, $\bfs_t$ is the scaling/geometry we use for clipping and noising. In Line~\ref{line:scale-then-clip}, rather than clip $\bfg_{t,j}$ we clip $\bfs_t \odot \bfg_{t,j}$. In Line~\ref{line:scaled-noise} we noise the scaled gradient, and then undo the scaling by $\bfs_t$ to return the clipped gradients to the original geometry before passing it to the \adam black-box. Since we are adding isotropic noise to the clipped objects, this has the same DP guarantees as \cref{fig:adam-pp}. An equivalent view is that we are clipping $\bfg_{t,j}$ to a non-isotropic ellipsoid defined by $\bfs_t$ and adding noise in the shape of this ellipsoid, where when $\hat{\bnu}$ is constant across coordinates, this ellipsoid is a sphere. \citet{li22side, li2023differentially} argued for scale-then-privatize (referring to it as clipping the preconditioned gradients) because it avoids adding excess noise in coordinates where the scaling $\bfs_t$ is small.

\mypar{Connection to delayed preconditioners} Scale-then-privatize is an extreme instantiation of the idea of delayed preconditioners of \cite{li2023differentially}. Delayed preconditioners is a technique where in each set of $s_1 + s_2$ iterations we (i) do $s_1$ iterations of doing no scaling of gradients and accumulating the average of the corresponding $s_1$ batches of gradients (ii) after $s_1$ iterations, use the gradient accumulated over those iterations to update $\hat{\nu}$, and then (iii) do $s_2$ iterations where we use $\hat{\nu}$ to scale the gradients ($\hat{\nu}$ is not updated in these iterations). Scale-then-privatize is equivalent to setting $s_1 = 0, s_2 = 1$. We do not extensively study delayed preconditioners separately from scale-then-privatize as (i) the use of both un-preconditioned and preconditioned gradients, as well as the reduced frequency of updates to $\hat{\nu}$, are cumbersome to include in our theoretical studies and (ii) the added hyperparameters $s_1, s_2$ need to be separately tuned which greatly increases the time needed to run experiments using delayed preconditioners. 

\subsubsection{Noising the Update}

Another private variant of \dpadam one could consider is to add isotropic noise to the updates $\hat{\bmu}/\sqrt{\hat{\bnu}}$ directly rather than to the gradients. For simplicity lets consider \adam with no momentum in the first moment (i.e., setting $\beta_1 = 0$) and $\epsilon_s = 0$. Now our goal is simply to noise the individual updates $\bfg_t / \sqrt{\hat{\bnu}_t}$. The main issue with this approach is that the worst-case sensitivity of the updates can be quite poor. Consider the one-dimensional setting, when the gradient of the first $t-1$ batches is 0, and the $t$-th batch $G_t$ has a single example with a non-zero gradient $\bfg_{t,j}$. Then, we will have $\hat{\bnu}_t = \frac{1-\beta_2}{1-\beta_2^t} \bfg_t^2$, so $\bfg_t / \sqrt{\hat{\bnu}_t} = \sign(\bfg_{t,j}) \cdot \sqrt{\frac{1-\beta_2^t}{1-\beta_2}}$. For $\beta_2 \approx 0.999$ this could be as large as $\approx 30$ for sufficiently large $t$, and furthermore it is not a function of the batch size, i.e. the sensitivity of the updates cannot be made to vanish by increasing the batch size. Even worse, this is only the sensitivity analysis when considering a single update. In reality, the gradient $\bfg_{t,j}$ will affect the value of $\hat{\bnu}_{t+1}, \hat{\bnu}_{t+2}, \ldots$, and hence indirectly affect the magnitude of all updates from $t$ to $T$, so the actual sensitivity of the sequence of all updates with respect to $\bfg_{t,j}$ could be much larger than the sensitivity of just the $t$-th update. In short, noising the updates requires too much noise to be practical unless we can arrive at a far more refined sensitivity analysis.

However, from the idea of noising the update we can arrive at scale-then-privatize; if we had $\hat{\bnu}_{t-1} = \hat{\bnu}_t$, for $\beta_1 = 0$ one can see that scale-then-privatize actually does add isotropic noise to the update (this is an idea we will revisit in \cref{sec:mf-pc}). Hence, one intuition for scale-then-privatize is that as long as $\hat{\bnu}$ does not change too much in any single iteration, scale-then-privatize approximately retrieves the idea of noising the updates, but with a far smaller amount of noise required.

\subsection{Preconditioner from Side Information}\label{sec:adam-side}

Finally, as a baseline one could consider computing a preconditioner $\bnu$ prior to training or during training using some auxiliary data, e.g. in-distribution public data, and then use this value rather than the noisy $\hat{\bnu}_t$ to choose per-coordinate learning rates. \citet{asi2021adapting, li22side} proposed the online version assuming some side information. In practice sources of side information such as in-distribution public data are  not readily available \cite{tramer2024considerations}, hence this method largely serves as a baseline rather than a practical method of interest in our setting.

\section{Theoretical Explorations}\label{sec:theory}

In this section, we make a number of claims about the different variants of \adam discussed in \cref{sec:variants}, for each trying to give some theoretical justification. At the end of each section we summarize the main takeaway in a highlighted text box, for readers who are interested in the claims moreso than the theoretical justifications for them.

\subsection{A Quadratic Example}\label{sec:quadratic}

To better understand the impact of noise on \dpadam and its variants, we introduce the following running example we will reference throughout the section: Consider a two-dimensional quadratic loss where the loss of model $(x, y)$ on example $(a, b)$ is $\frac{c_x}{2}(x-a)^2 + \frac{c_y}{2}(y-b)^2$. We assume the examples $(a, b)$ come from the distribution $N(\boldzero, \mathbb{I})$, and in each iteration we sample a new example from this distribution and use its gradient in the first-order optimizer.

If we start optimization with $x, y \approx 0$, for most gradients $(g_x, g_y)$ we sample we will have $|g_x|$ is within a constant factor of $c_x$, and $|g_y|$ is within a constant factor of $c_y$. For (non-private) \adam, this means we will have $\bnu \approx (c_x^2, c_y^2)$ and hence \adam would use per-coordinate learning rates $\approx (1/c_x, 1/c_y)$. For this axis-aligned loss, this approximately retrieves using the inverse of the Hessian as a scaling on the gradient, i.e. the updates used in Newton's method. In settings where $c_x \gg c_y$ or $c_x \ll c_y$, this will lead to much faster convergence than using the same learning rate for both coordinates.

What happens if we use \dpadam (via post-processing) instead? For now, let us ignore clipping and just consider \adam with a noise $(z_x, z_y) \sim N(\boldzero, \sigma^2 \mathbb{I})$ added to each gradient. The expected per-coordinate square of the gradient in each iteration is now $(c_x^2 + \sigma^2, c_y^2 + \sigma^2)$. In turn, if $\sigma \gg c_x$ or $\sigma \gg c_y$, we are going to use a smaller per-coordinate learning rate $\approx 1/\sigma$ in the corresponding coordinate. Put another way, one intuition for \dpadam is that the noise imposes a floor of $1/\sigma$ on the per-coordinate learning rates, which prevents \dpadam from distinguishing between coordinates where the ``correct'' per-coordinate learning rate is smaller than $1/\sigma$.

\subsection{Variance from Privacy vs. Variance from Data}
\label{sec:direction}

A seemingly key advantage of independent moment estimation (\cref{fig:adam-ime}) and bias correction (\cref{fig:adambc}) is that the second moment estimate $\hat{\bnu}_t$ is unbiased. However, using the quadratic example from \cref{sec:quadratic}, we argue that this is not the right goal for designing private adaptive optimizers.

Let's consider the following ``noiseless preconditioner'' variant of \dpadam: We use the noisy gradients to compute $\bmu$, but compute $\bnu$ using noiseless gradients. This is effectively \cref{fig:adam-ime} or \cref{fig:adambc} if they could achieve an even stronger goal of computing $\bnu$ exactly, rather than just exactly in expectation. Is this algorithm necessarily preferable over \dpadam via post-processing? While this algorithm uses the right learning rates $(1/c_x, 1/c_y)$ for the non-private distribution of gradients, for the private distribution of gradients these are not necessarily the right learning rates. One way to see this: Consider the (non-private) gradient distribution if instead of drawing the samples from $N(0, \mathbb{I})$ we draw the samples $(a, b)$ from 

\[N\left(\boldzero, 
\begin{bmatrix}
\frac{c_x^2 + \sigma^2}{c_x^2} & 0 \\
0 & \frac{c_y^2 + \sigma^2}{c_y^2} 
\end{bmatrix}
\right).\]

The gradient distribution for this data distribution is the same as the gradient distribution we get from original data distribution of $N(\boldzero, \mathbb{I})$ if we additionally add noise for privacy $N(0, \sigma^2 \mathbb{I})$ added to the gradients. However, for this gradient distribution non-private \adam will not try to learn the preconditioner $(c_x^2, c_y^2)$ which \cref{fig:adam-ime} and \cref{fig:adambc} are trying to learn, but the preconditioner which is learned by \dpadam via post-processing (\cref{fig:adam-pp}). This means one of two things is true: Either (i) ``noiseless preconditioner'' \dpadam (and algorithms like \cref{fig:adam-ime} and \cref{fig:adambc} which are trying to emulate it) is not the right algorithm to mimic: there are two learning problems which have the same gradient distributions, i.e. are fundamentally the same learning problem, but for which noiseless preconditioner \dpadam leads to wildly different optimization behaviors.

\begin{tcolorbox}[width=\textwidth,colback={SkyBlue}]    
   \textbf{Takeaway:} For quadratic loss functions, the gradient distribution with DP noise matches the gradient distribution for a different data distribution without DP noise. Algorithms like bias correction and independent moment estimation will arrive at different preconditioners for these two problems, even though they have the same gradient distribution. So an unbiased estimate of the preconditioner is unlikely to be the right goal for private adaptive optimizers.
\end{tcolorbox}    

\subsection{Benefits of Scale-Then-Privatize}

We argue that, at least in idealized settings, scale-then-privatize simultaneously: (i) adequately matches its preconditioner to the geometry of the non-private gradients (ii) matches the geometry of the non-private gradients and private gradients, such that property (i) is desirable even in light of the previous section.

\subsubsection{Scale-Then-Privatize Roughly Retrieves the Non-Private Geometry}\label{sec:retrievenonprivate}

Our first claim in this section is that, at least in idealized settings, scale-then-privatize arrives at a preconditioner roughly proportional to the preconditioner we'd arrive at non-privately, which by rescaling the learning rate allows us to retrieve some of the benefits of non-private adaptive optimizers. 

We will study the setting where $\bfg_t = \bfg$ in all rounds (and hence $\hat{\bnu}_t^* = \bfg^2$ in all rounds) and $\stability$ is negligible and can be treated as zero. While this is a very simple setting, (i) as demonstrated in the previous section, other approaches proposed in the literature fail in simple settings, so it is worth validating that an approach works in such settings and (ii) we believe this setting still gives good intuition for the desirable behaviors of scale-then-privatize. 

For ease of presentation, we fix $\clipnorm = 1, \beta_2 = 0.999$, the latter being a standard choice for \adam. We will also focus on the later iterations of training where $t$ is large, such that we can use the approximation $\beta_2^t \approx 0$. We give the same analysis in more generality in \cref{sec:retrievenonprivate-general}.

Using these settings and approximations, we start by rewriting $\hat{\bnu}_t$ in \cref{fig:adam-pc} as:

\begin{align}
\hat{\bnu}_t &= \bnu_t = 0.999 \cdot \bnu_{t-1} + 0.001 \cdot \tilde{\bfg}_t^2 = \nonumber\\
&=  0.999 \cdot \hat{\bnu}_{t-1} + 0.001\left(\bfg_t + \frac{1}{B}\left(\bfz_t / \bfs_t\right)\right)^2 = 0.999 \cdot \hat{\bnu}_{t-1} + 0.001 \left(\bfg_t + \frac{1}{B}\left(\bfz_t \odot \left(\sqrt{\hat{\bnu}_{t-1}} + \stability\right)\right)\right)^2\label{eq:hatnurecurrence-simple}
\end{align}

Taking expectation over $\bfz_t$ (conditioned on all randomness before iteration $t$):

\[\mathbb{E}[\hat{\bnu}_t] = 0.999 \cdot \hat{\bnu}_{t-1} + 0.001 \left(\bfg^2 + \frac{\sigma^2}{B^2} \hat{\bnu}_{t-1}\right) = \left(0.999 + 0.001 \cdot \frac{\sigma^2}{B^2} \right)\hat{\bnu}_{t-1} + 0.001 \cdot \bfg^2.\]

Now we have that if

\[\hat{\bnu}_{t-1} = \frac{1}{1 - \frac{\sigma^2}{B^2}} \bfg^2 = \frac{1}{1 - \frac{\sigma^2}{B^2}} (\hat{\bnu}_{t-1}^*)^2,\]

then $\mathbb{E}[\hat{\bnu}_t] = \hat{\bnu}_{t-1}$, i.e. $\hat{\bnu}$ does not change in expectation. In other words, assuming $\sigma / B < 1$, a ``steady state'' for $\hat{\bnu}$ (in expectation) is proportional to $\hat{\bnu}^*$ as desired. Furthermore, in coordinates where $\hat{\bnu}_{t-1} > \frac{1}{1 - \frac{\sigma^2}{B^2}} \bfg^2$, $\mathbb{E}[\hat{\bnu}_t] < \hat{\bnu}_{t-1}$ and similarly in coordinates where $\hat{\bnu}_{t-1} < \frac{1}{1 - \frac{\sigma^2}{B^2}} \bfg^2$, $\mathbb{E}[\hat{\bnu}_t] < \hat{\bnu}_{t-1}$, i.e. $\hat{\bnu}_t$ converges towards its steady state. In more detail, let's focus on a single coordinate of $\hat{\bnu}$ (i.e., the one-dimensional setting) and suppose $\hat{\bnu}_{t-1} = \frac{\alpha}{1 - \frac{\sigma^2}{B^2}} \bfg^2$, i.e. $\alpha$ times the steady state. Then:

\begin{align*}
\mathbb{E}[\hat{\bnu}_t] &= \left(0.999 + 0.001 \cdot \frac{\sigma^2}{B^2} \right)\frac{\alpha}{1 - \frac{\sigma^2}{B^2}} \cdot \bfg^2 + 0.001 \cdot \bfg^2\\
&= \left(1 + (\alpha - 1) \left(0.999 + 0.001 \cdot \frac{\sigma^2}{B^2}\right)\right) \cdot \frac{1}{1 - \frac{\sigma^2}{B^2}} \cdot \bfg^2\\
\end{align*}

So, in expectation, the additive error reduces from $(\alpha - 1)$ times the steady state to $(\alpha - 1)\left(0.999 + 0.001 \cdot \frac{\sigma^2}{B^2}\right)$ times the steady state. If $\frac{\sigma^2}{B^2} < 1$ this means in expectation $\hat{\bnu}_t$ will be closer to the steady state $\frac{1}{1 - \frac{\sigma^2}{B^2}}\bfg^2$ than $\hat{\bnu}_{t-1}$. In other words, when using \cref{fig:adam-pc} not only is the ``steady state'' of $\hat{\bnu}$ proportional to $\hat{\bnu}^*$, but $\hat{\bnu}$ will converge to $\hat{\bnu}^*$ in expectation. 

\begin{tcolorbox}[width=\textwidth,colback={SkyBlue}]    
   \textbf{Takeaway:} Scale-then-privatize (in expectation) retrieves a preconditioner proportional to the preconditioner arrived at by the noiseless gradients. This is effectively the same as using the noiseless preconditioner and rescaling the learning rate.
\end{tcolorbox}    

\subsubsection{Scale-Then-Privatize Aligns the Private and Non-Private Geometry}\label{sec:pcaligns}

We now return to the quadratic example of \cref{sec:quadratic} and see how scale-then-privatize affects the behavior in contrast with noiseless preconditioner \dpadam. Per the previous section, we expect that the preconditioner used by \dpadam with scale-then-privatize (\cref{fig:adam-pc}) is roughly proportional to $(1/c_x, 1/c_y)$. If we assume the preconditioner is equal to $\alpha \cdot (1/c_x, 1/c_y)$, the distribution of the noise added to gradients is $(c_x / \alpha, c_y /\alpha) \odot (z_x, z_y)$, $z_x, z_y \sim N(0, 1)$. So the overall distribution of the gradients when $x = 0, y = 0$ is 

\[N\left(\boldzero, (1 + \alpha^2)
\begin{bmatrix}
c_x^2 & 0 \\
0 & c_y^2
\end{bmatrix}
\right).
\]

Hence the preconditioner $\alpha \cdot (1/c_x, 1/c_y)$ actually matches the geometry of the gradients, up to a proportionality of $\frac{\alpha^2}{1+\alpha^2}$ (which can be corrected for with an appropriate choice of learning rate). This example demonstrates that when \dpadam with scale-then-privatize retrieves the noiseless preconditioner (up to some proportionality), it also retrieves the right second moments of the public distribution of gradients (up to some other proportionality). This is in contrast with \dpadam via post-processing which matches the private preconditioner with the private gradient distribution but doesn't match the noiseless preconditioner even up to proportionality, or noiseless preconditioner \dpadam which exactly uses the noiseless preconditioner but whose gradient distribution is a mismatch with this preconditioner.

\begin{tcolorbox}[width=\textwidth,colback={SkyBlue}]    
   \textbf{Takeaway:} Unlike the other variants of \dpadam, scale-then-privatize causes the noisy gradient distribution to have a similar geometry to the noiseless gradient distribution. In light of the quadratic example, this is important as otherwise retrieving the noiseless preconditioner may not be useful.
\end{tcolorbox}    

\subsection{Potentially Negative Estimates of $\bnu$}

Our next claim is that there are two regimes for techniques which use unbiased estimates of $\hat{\bnu}_t^*$, depending on the ratio $R$ of the average coordinate of $\hat{\bnu}_t^*$ to the noise multiplier $\sigma$. As previously discussed, techniques such as bias correction and independent moment estimation which estimate $\hat{\bnu}_t^*$ in an unbiased manner can potentially underestimate $\hat{\bnu}_t^*$, and even produce negative estimates. 
We first attempt to quantify when this is a concern, and then explain the potential adverse effects of having negative coordinates in $\hat{\bnu}^*$. 

\subsubsection{Regimes for Unbiased Estimators of $\bnu$}
\label{sec:regimes}

First, we try to more formally describe the regime when methods such as bias correction and independent moment estimation can produce $\bnu$ with many negative coordinates. We primarily consider bias correction, i.e. \cref{fig:adambc}. We will later discuss how the results would change for independent moment estimation (\cref{fig:adam-ime}).

\mypar{The one-dimensional case}
For simplicity let's fix $\clipnorm = 1$, assume we are in the one-dimensional case and $\bfg_t$ (the averaged clipped gradient) is equal to $\bfg$ for all $t$. We will again fix $\beta_2 = 0.999$; we state the inequalities in this section for general $\beta_2$ in \cref{sec:regimes-general}. 

Similarly to \cref{eq:hatnurecurrence-simple}, we have:

\[\hat{\bnu}_1 = \left(\bfg + \frac{1}{B} \bfz_1\right)^2, \qquad
\hat{\bnu}_t = \frac{0.999 - 0.999^t}{1 - 0.999^t}\hat{\bnu}_{t-1} + \frac{.001}{1-0.999^t}\left(\bfg + \frac{1}{B} \bfz_t\right)^2.\]

Unrolling the recursion we have:

\[\hat{\bnu}_t = \sum_{j=1}^t \frac{.001 \cdot 0.999^{t-j}}{1-0.999^t} \left(\bfg + \frac{1}{B}\bfz_j\right)^2 = \hat{\bnu}_t^* + \sum_{j=1}^t \frac{.001 \cdot 0.999^{t-j}}{1-0.999^t} \left(\frac{1}{B}\bfg \bfz_j + \frac{1}{B^2}\bfz_j^2\right).\]

Since $\bfz_t$ is a Gaussian and thus symmetric, $\Cov{\bfz_t}{\bfz_t^2} = 0$. So the variance of $\hat{\bnu}_t$ is

\begin{align*}
\Var{\hat{\bnu}_t} &= \left(\frac{1-0.999}{1-0.999^t}\right)^2 \sum_{j=1}^t 0.999^{2(t-j)} \Var{\frac{1}{B}\bfg \bfz_j + \frac{1}{B^2}\bfz_j^2} \\
&= \left(\frac{0.001}{1-0.999^t}\right)^2 \sum_{j=1}^t 0.999^{2(t-j)} \left(\frac{\bfg^2}{B^2}\Var{ \bfz_j} + \frac{1}{B^4}\Var{\bfz_j^2}\right)\\
&= \left(\frac{\bfg^2 \sigma^2}{B^2} + \frac{2 \sigma^4}{B^4}\right) \left(\frac{0.001}{1-0.999^t}\right)^2 \sum_{j=1}^t 0.999^{2(t-j)}\\
&= \left(\frac{\bfg^2 \sigma^2}{B^2} + \frac{2 \sigma^4}{B^4}\right) \left(\frac{0.001}{1-0.999^t}\right)^2 \frac{1 - 0.999^{2t}}{1 - 0.999^2}\\
&= \left(\frac{\bfg^2 \sigma^2}{B^2} + \frac{2 \sigma^4}{B^4}\right) \frac{0.001^2}{1 - 0.999^2} \cdot \frac{1 - 0.999^{2t}}{(1 - 0.999^t)^2} \\
&\geq \left(\frac{\bfg^2 \sigma^2}{B^2} + \frac{2 \sigma^4}{B^4}\right) \frac{1}{1999} \\
\end{align*}

In the last step we use the fact that $\frac{1-x^2}{(1-x)^2}$ is equal to 1 at $x = 0$ and increasing in $x \in [0, 1)$. 

As a technical aside, if we were using independent estimation, recall that each noise term in $\hat{\bnu}_t$ has standard deviation $\approx \clipnorm^2 \sigma / B$. The corresponding lower bound for $\Var{\hat{\bnu}_t}$ would by a similar derivation be $\frac{\sigma^2}{1999 B^2}$. Since $|\bfg| \leq 1$, the lower bound on $\Var{\hat{\bnu}_t}$ for bias correction is smaller, i.e. whatever conclusions we arrive at for the regime where bias correction is effective would be optimistic if applied to independent moment estimation.

Now, in order for the bias corrected estimate $\hat{\bnu}_t - \frac{\clipnorm^2 \sigma^2}{B^2}$ to be nonnegative with at least, say, probability $1/2$, a necessary condition is for the standard deviation to be at most a small constant factor times the magnitude of $E[\hat{\bnu}_t - \frac{\clipnorm^2 \sigma^2}{B^2}] = \bfg^2$. Rearranging, and using $x \lesssim y$ to denote $x \leq cy$ where $c$ is a small constant, we get:

\begin{align*}
\sqrt{\Var{\hat{\bnu}_t}} &\lesssim \bfg^2 \\
\Var{\hat{\bnu}_t} &\lesssim \bfg^4\\
\rightarrow \left(\frac{\bfg^2 \sigma^2}{B^2} + \frac{\sigma^4}{B^4}\right) \frac{1}{1999} &\lesssim \bfg^4
\\
\rightarrow B / \sigma &\gtrsim \frac{1}{1999^{1/4}} \frac{1}{|\bfg|} \approx 0.15 \cdot \frac{1}{|\bfg|}\\
\end{align*}

In the one-dimensional case we might (optimistically) expect $\bfg \approx 1$, in which case this condition is easily satisfied even for very small batch sizes $B \approx 0.15 \sigma$. Even for a more pessimistic $\bfg \approx 1 / \sqrt{B}$ (what we would get if each individual gradient $\bfg_{t,j}$ was sampled to be $1$ or $-1$ w.p. $1/2$ each), we need $\sqrt{B} \gtrsim 0.15 \sigma$ and this condition is still easily satisfied.

\mypar{The optimistic high-dimensional case} In the $\mdim$-dimensional setting, at least $1/2$ the coordinates of $\bfg_t$ (the average of the clipped gradients in round $t$) satisfy $\bfg_t(i) \le \sqrt{2} \ltwo{\bfg_t} / \sqrt{\mdim}$. So even if we optimistically assume $\ltwo{\bfg_t} = 1$ (which assumes all examples' gradients are the same)
the sufficient condition for, say, at most $1/4$ of the coordinates of $\bfg_t$ having negative $\hat{\bnu}$ with probability at least $1/2$ becomes:

\begin{equation}\label{eq:regimes}
B / \sigma \gtrsim 0.15 \sqrt{\mdim}.
\end{equation}

In other words, as long as the dimension $\mdim$ is sufficiently small, a method such as bias correction will not produce negative estimates of the second moment frequently. However, if the dimension $\mdim$ is sufficiently large as a function of the batch size $B$, bias correction will frequently produce negative estimates of the second moment, which even with the stability constant can lead to improperly large per-coordinate learning rates (or a large stability constant to regularize these learning rates) and in turn undesirable learning outcomes (we will discuss this further in the next section). As an example, consider setting $B = 2048, \sigma = 1$. In this setting, this condition says the dimension should be $\mdim \lesssim 10^6$ to avoid undesirable learning outcomes. Accounting for the fact that some small constants are masked by the approximate inequality, this condition might reasonably hold for research-scale models, but not for slightly larger ones.

\mypar{The pessimistic high-dimensional case} If $G_t$, the per-example gradients in the $t$-th batch, were not all the same but instead orthogonal, we would actually get $\ltwo{\bfg_t} \leq \clipnorm / \sqrt{B}$. If we instead plug in $\ltwo{\bfg} = \clipnorm / \sqrt{B\mdim}$ for the average coordinate value into \eqref{eq:regimes}, we get the condition:

\begin{equation}\label{eq:regimes-pessimistic}
\sqrt{B} / \sigma \gtrsim 0.15 \sqrt{\mdim}.
\end{equation}

This is now a much more stringent condition. Consider again setting $B = 2048, \sigma = 1$, then this condition says the dimension should be  $\mdim \lesssim 2 \cdot 10^4$, far smaller than any modern model size (e.g., even a CNN for MNIST classification has $5 \cdot 10^5$ parameters).

\mypar{The high-dimensional case with prefix sums} If using independent moment estimation, we could instead use \dpmf, i.e. noisy prefix sums, to add correlated noise to $\hat{\bnu}_t$ instead. This might improve the regime in which unbiased estimates of $\hat{\bnu}_t$ avoid frequent negative estimates. For a noisy prefix sum of $T$ terms $\sum_{t=1}^T x_t$ with equal sensitivity, \dpmf can achieve variance scaling as $O(\log^3 T)$ as opposed to $O(T)$ achieved by independent noise. Due to the decay parameter $\beta_2$, $\hat{\bnu}_t$ is not a prefix sum of equally-weighted terms, but a decaying prefix sum $\sum_{t=1}^T \beta_2^{T-t} x_t$. Determining the variance reduction for decaying prefix sums is a question that to the best of our knowledge has not been addressed by the literature. We can instead optimistically assume \dpmf's variance is that of just privatizing the ``highest-weight'' term $x_T$ in each decaying prefix sum (e.g., in the decay-less case, this corresponds to assuming the variance is $O(1)$ instead of $O(\log^3 T)$). By almost the same derivation, this gives the following lower bound on the variance for e.g. bias-correction for the setting $\clipnorm = 1, \beta_2 = 0.999$:

\[\Var{\hat{\bnu}_t} \geq 10^{-6} \left(\frac{\bfg^2 \sigma^2}{B^2} + \frac{2 \sigma^4}{B^4}\right).\]

This translates to, for example, the following bound on $B / \sigma$ analogous to e.g. \cref{eq:regimes} for the optimistic high-dimensional case:

\[ B / \sigma \gtrsim .032 \sqrt{\mdim}.\]

This is a $5 \times$ improvement over \eqref{eq:regimes-pessimistic}, i.e. a small constant improvement even with an optimistic assumption on the improvement in the variance. In short, this means that using correlated noise does not significantly increase the range of $B, \sigma, \mdim$ for which unbiased estimates of $\hat{\bnu}_t$ avoid bad learning outcomes.

\begin{tcolorbox}[width=\textwidth,colback={SkyBlue}]    
   \textbf{Takeaway:} In low-dimensional settings, even with small batch sizes bias correction and independent moment estimation are unlikely to produce negative preconditioner values. However, even for model dimension as small as that used by e.g. CNNs for MNIST classification, bias correction or independent moment estimation are at risk of producing many negative preconditioner values unless we use a batch size in the tens or hundreds of thousands, far larger than what is common in practice.
\end{tcolorbox}    

\subsubsection{Negative Effects of Negative $\bnu$}\label{sec:negativeeffects}

Next we explore why having unbiased estimators of $\bnu^*$, or equivalently many negative values in $\bnu$, may be harmful for learning. Recall that the default per-coordinate learning rate $1/\sqrt{\bnu}$, which is undefined or imaginary if $\bnu$ is non-positive. Hence, bias correction and independent moment estimation effectively use $1/\sqrt{\max\{\bnu, \stability^2\}}$ where $\stability$ is a stability constant instead, to ensure the per-coordinate learning rate is always well-defined and not too large.

What happens if the stability constant $\stability$ is very small? Where $\bnu$ is negative, we will end up with a per-coordinate learning rate of $1/\stability$ which is very large. For non-private \adam when the stability constant is the dominating term in $\bnu$ and the per-coordinate learning rate is $\approx 1 / \stability$ this is okay, because this means the gradient in the corresponding coordinate has magnitude at most some constant times $\stability$, so the resulting update is at most constant in size.

In the presence of noise however, this large per-coordinate learning rate can be much more problematic, at least for the methods we survey: For both bias correction and independent moment estimation, it is possible to have a negative coordinate in $\bnu$, but for the corresponding coordinate of the (noisy) gradient to have a magnitude much larger than $\stability$, either due to the bias correction term or due to the independent noises in independent moment estimation. As a result the update for this coordinate will be very large. If this happens for many coordinates simultaneously, it could potentially cause training to complete diverge. \citet{kalinin2025continual} propose to clip the final update to avoid this destabilization, but even with this mitigation, the clipped update will have most of its magnitude on coordinates where this destabilization occurred, i.e. the resulting optimizer will be slow to make progress on coordinates which did not destabilize. 

We can of course mitigate this by choosing $\stability$ to be larger. However, choosing $\stability$ to be larger will homogenize the values of $\bnu$ in coordinates where $\bnu^*$ is smaller. The bias of the post-processing approach is considered harmful because of this homogenization, so using too large a value of $\stability$ defeats the point of using an unbiased estimator of $\bnu^*$ in the first place. 

In \cref{sec:regimes} we showed \dpadam with bias correction the standard deviation of $\bnu_i$ is at least a constant times $\frac{\clipnorm^2 \sigma^2}{B^2}$, the bias of $\bnu$ in \dpadam via post-processing. So if we are in a regime where a constant fraction of coordinates of $\bnu$ have magnitude at most $\frac{\clipnorm^2 \sigma^2}{B^2}$, i.e. we are in a regime where training is likely to destabilize for small $\stability$, then we need to set $\stability^2$ to be about as large as the bias correction term $\frac{\clipnorm^2 \sigma^2}{B^2}$ to ensure that the magnitude of the update in any coordinate is at most a constant. But if we set $\stability^2$ to be as large as the bias correction term, we effectively add the bias correction term back to the smallest coordinates of $\bnu$, undoing the benefits of bias correction. In particular, these are the coordinates where the gradient signal is sparsest or smallest, and where we arguably could benefit the most from adaptive optimization.

For independent moment estimation the high-level story is similar, although the negative values in $\bnu$ now come from adding negative noise to $\bnu$ rather than from subtracting a bias correction. This introduces an additional adverse effect, where the per-coordinate learning rate of a coordinate could be $1/\stability$, but the gradient could have a large noise value added to it, and we will hence take a large noise update on this coordinate. Put another way, \dpadam via post-processing (or with bias correction) has a self-regularization behavior where if large noise is added to one coordinate of the gradient, the corresponding coordinate of $\bnu$ will be large, hence the per-coordinate learning rate will be small and we avoid taking a large noisy update in that coordinate. The decoupling of noises in independent moment estimation prevents this self-regularization.

In short, in the regime where many of $\bnu^*$'s coordinates are smaller than the bias $\frac{\clipnorm^2 \sigma^2}{B^2}$, \dpadam with bias correction or independent moment estimation is forced to either destabilize, clip the updates and progress slowly on coordinates where $\bnu$ is large, or use a large stability constant that effectively undoes the bias correction. We emphasize that these behaviors may not be problematic if most coordinates of $\bnu$ are sufficiently large, which e.g. is more likely to happen in low-dimensional settings. Hence our claim is not that bias correction or independent moment estimation universally introduce adverse effects to private adaptive optimizers, just that there are some parameter regimes where they do.

\begin{tcolorbox}[width=\textwidth,colback={SkyBlue}]    
   \textbf{Takeaway:} When bias correction or independent moment estimation produce negative preconditioner values, the ways to correct for them either destabilize learning, require heavily slowing down the learning of \adam, or require giving up on the benefits of adaptivity. Hence these variants of \dpadam may be undesirable in the regimes where they produce many negative preconditioner values.
\end{tcolorbox}    

\mypar{Conclusions} To summarize this section, one can think of there being two regimes in terms of $\mdim, B, \sigma$. One regime is where \cref{eq:regimes} (or \cref{eq:regimes-pessimistic}, if being pessimistic) holds, and methods such as bias correction avoid undesirable learning outcomes (and as we will demonstrate later, may even be strongly preferable to other methods). The second regime is where \cref{eq:regimes} fails to hold, and bias correction may actually impede learning performance.

Based on these observations, we caution against the use of DP variants of \adam which preserve unbiasedness of the second moment estimate in high-dimensional regimes, and conclusions from experiments using these variants in low-dimensional settings should not be expected to hold in higher-dimensional settings. 

\subsection{Scale-Then-Privatize and Prefix Sums}\label{sec:mf-pc}

We now consider the interplay between scale-then-privatize and \dpmf. The main goal of \dpmf is to cancel out noise across different iterations. When the underlying optimizer is \sgd, the learning rate remains constant across iterations so the total noise added to the sum of the first $t$ optimizer updates is a prefix sum proportional to $\sum_{j=1}^t (\Cinv\bfz)_j$. Hence, the line of work on \dpmf proposes to choose $\Cinv$ that minimizes the RMSE, i.e. average of the variances of prefix sums $\{\sum_{j=1}^t (\Cinv\bfz)_j\}_t$. This is equivalent to optimizing the tradeoff between noise cancellation and increased sensitivity, and can be done entirely offline, i.e. before starting training. 

Before understanding how to optimize $\Cinv$ for \adam, as an intermediate we can understand \sgd with momentum parameter $\beta_1$. \sgd with momentum initializes a momentum $\mu_0$, and uses the update $\mu_t = \beta_1 \mu_{t-1} + \bfg_t$, $\bftheta_t = \bftheta_{t-1} - \eta \mu_t$. 
Without clipping or noise, the iterates $\bftheta_{t+1}$ of \sgd with momentum $\beta_1$ can be expressed as a linear combination of the gradients $\bfg_1, \dots, \bfg_{t}$, namely
\[
\bftheta_{t+1} = \bftheta_0 + \sum_{j=1}^t \left(\sum_{k=0}^{t-j} \beta_1^k \right) \bfg_j,
\]
and hence the total noise in $\bftheta_{t+1}$ in \sgd with momentum using \dpmf to add noise is
\[
\sum_{j=1}^t \left(\sum_{k=0}^{t-j} \beta_1^k \right)  (\Cinv\bfz)_i 
\]
and we could again choose $\Cinv$ that minimizes the average variance of the noises $\{\sum_{j=1}^t \left(\sum_{k=0}^{t-j} \beta_1^k \right)  (\Cinv\bfz)_i \}$ before starting optimization. However, \citet{choquette2022multi} observe optimizing for this objective instead of the momentum-less objective has minimal to no benefit for training models; that is, we can use $\Cinv$ optimized for \sgd without momentum, even for \sgd with momentum, and empirically this is about as good. 

Now, we consider how to privatize \dpadam. The main challenge in contrast with \sgd is that in Algorithm~\ref{fig:adam-pp} (\dpadam via post-processing) and most other variants we've discussed, the noise added to the sum of the optimizer updates is (setting the stability parameter $\stability = 0$) proportional to $\sum_{j=1}^t  \sum_{k=j}^t \frac{\beta_1^{t-k}}{1 - \beta_1^t} \frac{(\Cinv\bfz)_j}{\sqrt{\hat{\bnu}_k}}$; note that up to the scaling by $\hat{\bnu}$ and normalization of the momentum $\frac{1}{1-\beta_1^t}$, this retrieves \sgd with momentum $\beta_1$. So if we assume that, similarly to \sgd, it suffices in practice to optimize for the momentum-less objective, then we want to minimize the average variance of the noises added to prefix sums $\{\sum_{j=1}^t   \frac{(\Cinv\bfz)_j}{\sqrt{\hat{\bnu}_j}}\}$. 

The main challenge is that even with this simplifying assumption, the term $\hat{\bnu}$ is determined in an online manner, and is not known prior to beginning optimization. That is, unlike \sgd, we cannot optimize the matrix $\Cinv$ in an offline manner because the coefficient $\hat{\bnu}$ in the prefix sum only appears during training. Furthermore, for \sgd the noise being added to each coordinate is the same, i.e. we can optimize a single $\Cinv$ and use it to noise all coordinates simultaneously. In contrast, for \dpadam even if $\hat{\bnu}$ was given offline, we would want to apply different choices of noise correlation $\Cinv$ to different coordinates, but e.g. DP composition theorems don't give us a straightforward way to do this. We argue that scale-then-privatize mitigates both these issues. For scale-then-privatize (\cref{fig:adam-pc}) we instead want to minimize the variance of:

\[\sum_{j=1}^t\frac{(\Cinv\bfz)_j / \bfs_j}{\sqrt{\hat{\bnu}_j}} = \sum_{j=1}^t\frac{(\Cinv\bfz)_j}{\sqrt{\hat{\bnu}_j / \hat{\bnu}_{j-1}}}.\]
Recall that 
\[\hat{\bnu}_t = \frac{\beta_2 - \beta_2^t}{1 - \beta_2^t}\hat{\bnu}_{t-1} + \frac{1-\beta_2}{1-\beta_2^t}\left(\bfg + \frac{\clipnorm}{B} (\Cinv\bfz)_t\right)^2,\]
and as $t$ increases $\frac{\beta_2 - \beta_2^t}{1 - \beta_2^t}$ approaches $\beta_2$ which is usually set to a constant close to 1 like $.999$ in practice. This means that without momentum, even if the squared gradients wildly differ in consecutive iterations, the scaling $\sqrt{\hat{\bnu}_j / \hat{\bnu}_{j-1}}$ approaches nearly $\boldone$. This mitigates both of the aforementioned issues: First, for optimizing $\bfC$ for \dpadam with scale-then-privatize, if we treat $\sqrt{\hat{\bnu}_j / \hat{\bnu}_{j-1}} = 1$ we retrieve an objective that can be optimized offline. Furthermore, this objective is nearly the same as that of \dpsgd with momentum, i.e. we can use the same methods to choose $\Cinv$ as for \dpsgd and be reasonably confident in the resulting choice of $\Cinv$ having good utility). Second, if $\sqrt{\hat{\bnu}_j / \hat{\bnu}_{j-1}} \approx 1$ then the objective is roughly the same for all coordinates, i.e. we can optimize a single $\Cinv$ and use it to noise all coordinates at a minimal loss in utility.

\begin{tcolorbox}[width=\textwidth,colback={SkyBlue}]    
   \textbf{Takeaway:} Without scale-then-privatize, designing correlated noise mechanisms for adaptive optimizers is challenging because we cannot optimize the mechanisms in an offline manner, and we may want to use different correlation strategies in different coordinates. In contrast, \dpadam with scale-then-privatize permits an offline objective that reasonably approximates the true objective, and also permits using the same correlation strategy across different coordinates at minimal loss in utility.
\end{tcolorbox}    

\section{Empirical Comparisons}\label{sec:empirical}

In this section we compare different variants of \dpadam, with and without \dpmf, in different settings.

\subsection{1-Dimensional Logistic Regression}

We first consider \adagrad (\cref{fig:adagrad}) in the 1-dimensional setting. We consider a synthetic sparse logistic regression setup. We generate data inputs $(x, y)$ where $x \sim N(0, 1), y \sim Bern(\frac{1}{1 + e^{-x}})$. Given an input $(x, y)$, $y \in \{0, 1\}$ and model $\bftheta$ the loss is $\ell(\bftheta; (x, y)) = y \log \left(\frac{1}{p}\right) + (1-y) \log \left(\frac{1}{1-p}\right)$ where $p = \frac{1}{1 + e^{-\bftheta x}}$; the ground truth corresponds to $\bftheta = 1$. We consider a sparse setting where for $90\%$ of the training data we set $x = 0$, i.e. the gradient $\nabla_{\bftheta} \ell(\bftheta; (x; y))$ is $0$. In this setting, using \adagrad (whose learning rate decays with the number of non-zero samples seen so far, instead of the number of samples seen so far) is preferable to \sgd with decaying learning rate. We generate 1000 training samples and do a single epoch with batch size 1, and use clip norm $\clipnorm = 1.0$ and noise multiplier $\sigma = 0.1$; in other words, we have $B / \sigma = 10$ which is a relatively small signal-to-noise ratio compared to modern large-scale DP training pipelnies. We tune the base learning rate $\eta$.

We compare four approaches, using \dpmf optimized for single-epoch training without any constraint on the strategy matrix $\bfC$:
\begin{itemize}
    \item The ground truth, i.e. $\bftheta = 1$.
    \item \adagrad without any noise.
    \item DP \adagrad via post-processing, i.e. we privatize the gradients and then use \adagrad as a black-box.
    \item DP \adagrad via independent moment estimation, i.e. we separately noise the gradient $\bfg_t$ and its square $\bfg_t^2$, and use the noisy $\bfg_t^2$ to update the noisy estimate $\hat{\bnu}$ of $\sum_{j=1}^t \bfg_j^2$. Since this estimate can be negative with noise, we use a learning rate of $\eta / \max\{1, \sqrt{\hat{\bnu}}\}$ instead of $\eta /  \sqrt{\hat{\bnu}}$  in round $t$. By default we use $\sigma = 0.1 \cdot \sqrt{2}$, but we also try independent moment estimation ``for free,'' i.e. using $\sigma = 0.1$.
\end{itemize}

\begin{figure}
\begin{center}
\begin{tabular}{|c|c|} 
 \hline
 Approach & Test Loss \\ [0.5ex] 
 \hline\hline
 Ground Truth ($\bftheta = 1$) & 0.5976  \\ 
 \hline
 Nonprivate \adagrad & 0.5976 \\
 \hline
 \adagrad via Post-Processing & 0.6010 \\
 \hline
 \adagrad via Independent Moment Estimation & 0.5981 \\
 \hline
 \adagrad via Independent Moment Estimation For Free & 0.5980 \\
 \hline
\end{tabular}
\end{center}
    \caption{Test loss of different variants of private \adagrad on 1-D logistic regression.}
    \label{fig:1dlr}
\end{figure}

In \cref{fig:1dlr} we compare the test loss of the final model on $10,000$ samples (not sparsified) of each approach. For the noisy variants, we report the average of 30 trials. We see that despite the relatively small value of $B / \sigma$, and even if independent moment estimation incurs extra noise, independent moment estimation achieves better performance than post-processing. While our later experiments will show that independent moment estimation is uncompetitive in the high-dimensional regime (matching the intuition built in Section~\ref{sec:theory}), this makeshift experiment provides evidence that in the low-dimensional regime it is a well-performing variant.

\subsection{High-Dimensional Transformer Training}

We next focus on a transformer training setup. We fine-tune a pretrained TinyBERT model (approximately $4.4$ million trainable parameters) on the masked token prediction task as defined in \cite{devlin2019bert} for the arXiv dataset. The arXiv dataset consists of $\approx$1.9 million papers. We use each paper's abstract, truncated to 512 tokens, as a text example for training. We randomly split the dataset into a train/test/validation split of $\approx$ 1.7/0.1/0.1 million examples each. We use 2000 iterations of training with a batch size of 2048 (such that each example participates at most $3$ times). We fix the clip norm of $1.0$ (when not using scale-then-privatize), which we found to be effective across all settings. As we are primarily concerned with comparing different variants of adaptive optimizers, which for a given DP guarantee will all use the same distribution of noise $\bfz_t$, we fix a noise multiplier across all experiments and ignore its mapping to a specific DP guarantee (including e.g. accounting for amplification, group sizes, etc.). For correlated noise, we use \bandmf of \cite{choquette2024amplified} and noise multiplier of $1$ (with $b =$128 bands for efficiency). We also use independent noise with noise multiplier of $\approx .134$ and $1$, such that the RMSE of prefix sums (as defined in e.g., \cite{choquette2022multi}) and the example-level $\epsilon$ respectively are the same for independent noise and correlated noise. We are primarily concerned about the effects of noise on \dpadam and less concerned about the specific $\epsilon$ the noise corresponds to, so we do not worry about amplification when computing $\epsilon$ for simplicity, i.e. our results on \dpmf vs \dpsgd do not necessarily indicate one or the other is better when combined with adaptive optimizers in settings where amplification is available.

In \cref{fig:tinybert} we compare the average test loss across 30 trials of the following variants of \dpadam:
\begin{itemize}
    \item Post-processing (\cref{fig:adam-pp}). We tune the learning rate only.
    \item Independent moment estimation (\cref{fig:adam-ime}). We tune the learning rate and the stability constant $\stability$.
    \item Bias correction (\cref{fig:adambc}). We tune the learning rate and the stability constant $\stability$.
    \item Scale-then-privatize (\cref{fig:adam-pc}). We tune the learning rate, stability constant used for clipping $\epsilon_{s_1}$, and re-tune the clip norm since we are no longer clipping in the $\ell_2$-norm.
    \item Noiseless preconditioner (a variant of \cref{fig:adam-side}). As previously discussed, we add noise to the gradients but use the value of $\bnu$ computed using the noiseless gradients. This is a non-private baseline purely for the sake of understanding the value of getting better estimates of $\bnu$. We try the noiseless preconditioner with and without scale-then-privatize.
\end{itemize}

We also validate the necessity of using \adam for this task by comparing to \sgd (with a tuned momentum parameter), both with and without noise.

\begin{figure}
\begin{center}
\begin{tabular}{llccc} 
 \toprule
 \multicolumn{2}{l}{Noising Mechanism} & \dpsgd, $\sigma = 0.134$ & \bandmf, $\sigma = 1.0$, & \dpsgd, $\sigma = 1.0$\\
 \multicolumn{2}{l}{} & & $b = 128$ & \\
 \midrule
 \multicolumn{2}{l}{RMSE of Prefix Sums} & $2.97 \cdot 10^{-3}$ & $2.97 \cdot 10^{-3}$ & $2.22 \cdot 10^{-2}$ \\
 \multicolumn{2}{l}{$\epsilon$ for $\delta = 10^{-7}$ (No Amplification)} & $149.1$ & $10.0$ & $10.0$\\
 \midrule
 \multicolumn{2}{l}{No fine-tuning} & & $4.955$ & \\
 \multicolumn{2}{l}{Nonprivate \sgd} & & $3.269$ & \\
 \multicolumn{2}{l}{Nonprivate \adam} & & $2.781$ &  \\ 
 \midrule
 \multirow{3}{*}{\dpadam} & Noiseless Preconditioner & $3.251$ & $3.497$ & $3.697$ \\
 & Scale-Then-Privatize &  \multirow{2}{*}{$3.071$} &  \multirow{2}{*}{$3.302$} &  \multirow{2}{*}{$3.659$} \\
 & + Noiseless Preconditioner & & & \\
 \midrule
 \multirow{4}{*}{\dpadam} & Post-Processing & $3.251$ & $3.493$ & $3.697$ \\
 & Independent Moment Estimation & $3.433$ & $3.520$ & $3.823$ \\
 & Bias Correction & $3.223$ & $3.492$ & $3.727$ \\
 & Scale-Then-Privatize & $\mathbf{3.097}$ & $\mathbf{3.423}$ & $\mathbf{3.659}$\\
 \midrule
 \multicolumn{2}{l}{\dpsgd} & $3.373$ & $3.605$ & $3.726$ \\
 \bottomrule
\end{tabular}
\end{center}
    \caption{Test loss of different variants of private \adam for fine-tuning TinyBERT. For all results except Bias Correction, the sample standard deviation is at most 0.001, and for Bias Correction it is at most 0.01.}
    \label{fig:tinybert}
\end{figure}

Our main observation is that independent moment estimation consistently does worse than the baseline \dpadam via post-processing, and bias correction slightly improves on this baseline in a very high-epsilon setting but otherwise does worse. In contrast, scale-then-privatize consistently improves on the post-processing approach. This is in contrast with the 1-dimensional setting, i.e., the performance of independent moment estimation and bias correction degrades in higher dimensions, matching the intuition developed in \cref{sec:theory}. 

In addition to scale-then-privatize outperforming the other approaches, our noiseless preconditioner baselines offer some support for our theoretical observations. Without scale-then-privatize, switching to a noiseless preconditioner in \dpadam leads to roughly the same test loss. This suggests that even if we can retrieve the ability of \adam to learn the geometry of the noiseless of the gradients, the benefits of this are negated by the fact that this geometry is a mismatch with the geometry of the noisy gradients, as demonstrated with the running example of \cref{sec:quadratic}. In contrast, with scale-then-privatize, using a noiseless preconditioner improves utility, slightly for \dpsgd in the low-noise setting and significantly for \dpmf. This demonstrates that when the geometry of the noiseless and noisy gradients are similar (as we demonstrated in \cref{sec:pcaligns}), better estimates of the preconditioner are indeed helpful. 

We can further validate the theoretical intuition of \cref{sec:regimes} by demonstrating that bias correction retrieves many negative coordinates in this setting. In \cref{fig:negativecoordinates} we plot the fraction of coordinates of $\hat{\bnu}$ which are negative after bias correction, when using \dpadam with independent noise and $\sigma \approx 0.134$. We see that anywhere from $45\%$ to $50\%$ of coordinates are negative in most rounds of training. In \cref{fig:norm} we also check the norm of the averaged clipped gradients $\bfg_t$, divided by the clip norm (in our experiments, just $1.0$) for \dpadam via post-processing with independent noise and $\sigma \approx .134$. A value of 1.0 corresponds to the ``optimistic'' setting in \cref{sec:regimes}, and a value of $1/\sqrt{B} = 1/\sqrt{2048} \approx .022$ corresponds to the ``pessimistic'' setting. The average value across all rounds is roughly $0.27$, suggesting we are closer to the optimistic regime, i.e. most pairs of gradients have a reasonably high dot product with each other. The fact that the fraction of negative coordinates in \cref{fig:negativecoordinates} is large in spite of this can be explained by the magnitudes of the gradients across different coordinates. The analysis in \cref{sec:regimes} is optimistic in that it gives sufficient lower bounds for $B$ (as a function of $\sigma, \mdim$) assuming the magnitudes of gradients (and hence the values of $\hat{\bnu}$) are roughly uniform across the coordinates. In \cref{fig:nu_vs_noise} we plot the values of $\bnu$ arrived at by non-private \adam and compare them to the bias correction term $\frac{\clipnorm^2 \sigma^2}{B^2}$, and we see most coordinates fall below the bias correction term in magnitude. Furthermore, for bias correction and independent moment estimation, in all settings the best value of the stability constant $\epsilon_s$ was $10^{-4}$, which corresponds to imposing a floor on the preconditioner values of $10^{-8}$. As we can see in \cref{fig:nu_vs_noise} this exceeds the magnitude of the bias correction; this validates the intuition from \cref{sec:regimes} that in order to mitigate the negative effects of bias correction or independent moment estimation we need to use a stability constant sufficiently large to effectively re-introduce the bias of \dpadam via post-processing for small coordinates of $\hat{\nu}$.

\begin{minipage}[t]{.48\textwidth}
\begin{figure}[H]
    \centering
    \includegraphics[width=0.99\textwidth]{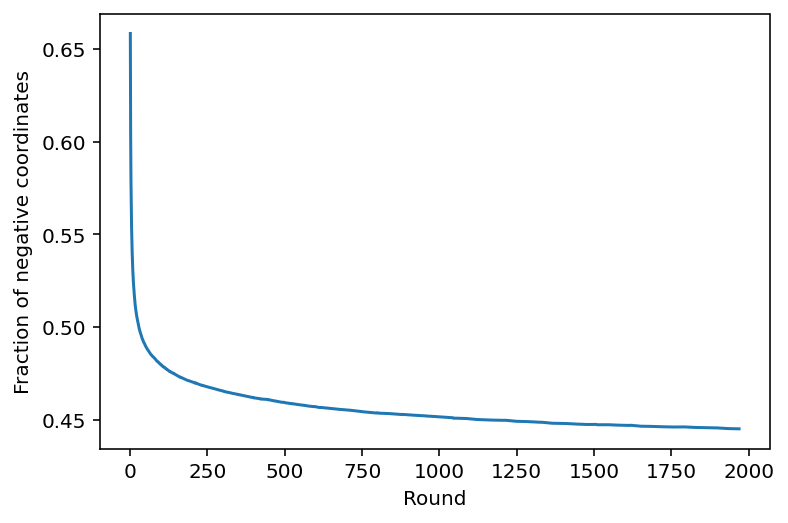}
    \caption{Fraction of coordinates which are negative in $\hat{\bnu}$ after bias correction.}
    \label{fig:negativecoordinates}
\end{figure}
\end{minipage}
\begin{minipage}[t]{.48\textwidth}
\begin{figure}[H]
    \centering
    \includegraphics[width=0.99\textwidth]{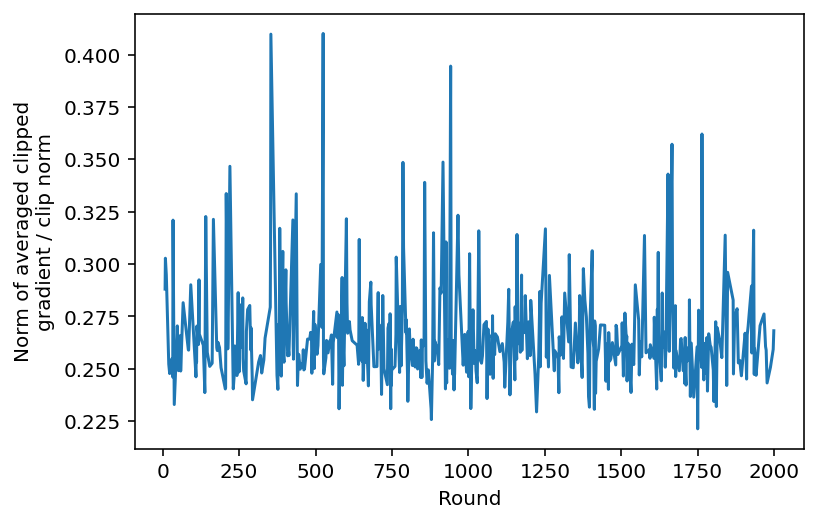}
    \caption{The norm of the averaged clipped gradient for \dpadam via post-processing.}
    \label{fig:norm}
\end{figure}
\end{minipage}
\begin{minipage}[t]{.48\textwidth}
\begin{figure}[H]
    \centering
    \includegraphics[width=0.99\textwidth]{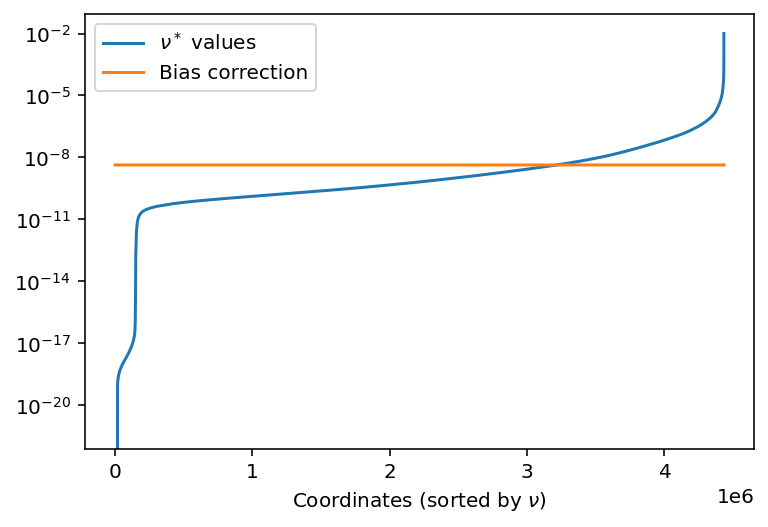}
    \caption{The values of non-private $\bnu$ compared to the bias correction term.}
    \label{fig:nu_vs_noise}
\end{figure}
\end{minipage}

Finally, we note that unlike what was observed in past works for \dpsgd, for \dpadam the RMSE of the noising mechanism is not a good predictor of learning performance, in spite of the observations made in \cref{sec:mf-pc}. This matches an observation made in \cite{mckenna2025scalingbandedmatrixfactorization}. It is an interesting future direction to explore what objectives for optimizing the choice of $\Cinv$ are best for adaptive optimizers.

\section*{Acknowledgements}

We are thankful to Ryan McKenna and Nikita Kalinin for their comments and corrections on an early draft of this manuscript.

\bibliographystyle{plainnat}
\bibliography{ref}

\appendix

\section{Bias Correction for \dpmf}\label{sec:dpmf-bc}

We focus on the one-dimensional case, i.e. the gradients are scalars, as in the higher-dimensional case the bias is just the bias in the one-dimensional case times $\boldone$. For \dpadam with \dpmf the second moment estimate in iteration $t$ is (given decay parameter $\beta < 1$):

\[\hat{\bnu}_t = \frac{1 - \beta_2}{1 - \beta_2^t} \cdot \sum_{i=1}^t \bfg_i^2 \cdot \beta_2^{i-1}.\]

Let $\bfD(\beta_2, t)$ be the diagonal matrix $\frac{1 - \beta_2}{1 - \beta_2^t} \cdot \diag(\beta_2^{t-1}, \beta_2^{t-2}, \ldots, \beta_2, 1)$. Then $\hat{\bnu}$ be written as $ \bfg_{:t}^\top \bfD(\beta_2, t) \bfg_{:t}$, where $\bfg_{:t} = (\bfg_1, \bfg_2, \ldots, \bfg_t)$. Defining $\bfz_{:t}$ analogously, the bias is then:

\begin{align*}
\mathbb{E}[(\bfg_{:t} + (\bfC^{-1}_{:t, :t}) \bfz_{:t})^\top \bfD(\beta_2, t) (\bfg_{:t} + (\bfC^{-1}_{:t, :t}) \bfz_{:t})] - \bfg_{:t}^\top \bfD(\beta_2, t) \bfg_{:t} &= \mathbb{E}[\bfz_{:t}^\top (\bfC^{-1}_{:t, :t})^\top \bfD(\beta_2, t) (\bfC^{-1}_{:t, :t}) \bfz_{:t}]\\
&= \Tr((\bfC^{-1}_{:t, :t})^\top \bfD(\beta_2, t)(\bfC^{-1}_{:t, :t})) \cdot \frac{\clipnorm^2 \sigma^2}{B^2}.
\end{align*}

We can also write this as a recursive formula. If the bias correction is $b_i$ in iteration $t$,  $b_0 = 0$, we have:

\[b_{t+1} = \frac{\beta_2 - \beta_2^{t+1}}{1-\beta_2^{t+1}} b_i +  \beta_2 \cdot \frac{1 - \beta_2}{1 - \beta_2^{t+1}} ((\bfC^{-1})^\top \bfC^{-1})_{t+1, t+1} \cdot \frac{\clipnorm^2 \sigma^2}{B^2}.\]

\section{Generalized Analysis from \cref{sec:retrievenonprivate}}\label{sec:retrievenonprivate-general}

We start by rewriting $\hat{\bnu}_t$ in \cref{fig:adam-pc} as:

\begin{align}
\hat{\bnu}_t &= \bnu_t / \left(1 - \beta_2^t\right) = \left(\beta_2 \bnu_{t-1} + \left(1 - \beta_2\right)\tilde{\bfg}_t^2\right) / \left(1 - \beta_2^t\right) = \frac{\beta_2 - \beta_2^t}{1 - \beta_2^t}\hat{\bnu}_{t-1} + \frac{1-\beta_2}{1-\beta_2^t}\tilde{\bfg}_t^2\nonumber\\
&=  \frac{\beta_2 - \beta_2^t}{1 - \beta_2^t}\hat{\bnu}_{t-1} + \frac{1-\beta_2}{1-\beta_2^t}\left(\bfg_t + \frac{\clipnorm}{B}\left(\bfz_t / \bfs_t\right)\right)^2 = \frac{\beta_2 - \beta_2^t}{1 - \beta_2^t}\hat{\bnu}_{t-1} + \frac{1-\beta_2}{1-\beta_2^t}\left(\bfg_t + \frac{\clipnorm}{B}\left(\bfz_t \odot \left(\sqrt{\hat{\bnu}_{t-1}} + \stability\right)\right)\right)^2\label{eq:hatnurecurrence}
\end{align}

Taking expectation over $\bfz_t$, conditioned on all the randomness before iteration $t$:

\[\mathbb{E}[\hat{\bnu}_t] = \frac{\beta_2 - \beta_2^t}{1 - \beta_2^t}\hat{\bnu}_{t-1} + \frac{1-\beta_2}{1-\beta_2^t}\left(\bfg^2 + \frac{\clipnorm^2 \sigma^2}{B^2} \hat{\bnu}_{t-1}\right) = \left(\frac{\beta_2 - \beta_2^t}{1 - \beta_2^t} + \frac{\clipnorm^2 \sigma^2}{B^2} \cdot \frac{1-\beta_2}{1-\beta_2^t}\right)\hat{\bnu}_{t-1} + \frac{1-\beta_2}{1-\beta_2^t} \bfg^2.\]

Let's assume $t$ is sufficiently large such that $\beta_2^t$ can be approximated as 0. Then letting $c = 1 + \frac{\clipnorm^2 \sigma^2}{B^2} \cdot \frac{1 - \beta_2}{\beta_2}$, we can rewrite the above more succinctly as:

\begin{equation}\label{eq:evolutionofnu}
\mathbb{E}[\hat{\bnu}_t] = c \beta_2 \hat{\bnu}_{t-1} + (1 - \beta_2) \bfg^2.
\end{equation} 

Now if we plug

\[\hat{\bnu}_{t-1} = \frac{1 - \beta_2}{1 - c \beta_2} \bfg^2 = \frac{1 - \beta_2}{1 - c \beta_2} \hat{\bnu}_{t-1}^*\]

into \cref{eq:evolutionofnu}, then we get $\mathbb{E}[\hat{\bnu}_t] = \hat{\bnu}_{t-1}$, i.e. $\hat{\bnu}$ does not change in expectation. In other words, assuming $c \beta_2 < 1$ (which holds for all sufficiently large batch sizes $B$, because $c \rightarrow 1$ as $B \rightarrow \infty$ and $\beta_2 < 1$), a ``steady state'' for $\hat{\bnu}$ (in expectation) is proportional to $\hat{\bnu}^*$ as desired. Furthermore, from \cref{eq:evolutionofnu}, we can observe that in coordinates where  $\hat{\bnu}_{t-1} > \frac{1 - \beta_2}{1 - c \beta_2} \hat{\bnu}_{t-1}^*$, $\mathbb{E}[\hat{\bnu}_t] < \hat{\bnu}_{t-1}$ and similarly in coordinates where $\hat{\bnu}_{t-1} < \frac{1 - \beta_2}{1 - c \beta_2} \hat{\bnu}_{t-1}^*$, $\mathbb{E}[\hat{\bnu}_t] > \hat{\bnu}_{t-1}$. In more detail, let's focus on the one-dimensional setting and suppose $\hat{\bnu}_{t-1} = \frac{\alpha (1 - \beta_2)}{1 - c \beta_2} \bfg^2$, i.e. $\alpha$ times the steady state. Then:

\begin{align*}
\mathbb{E}[\hat{\bnu}_t] &= \left(\frac{\alpha (1 - \beta_2) c \beta_2}{1 - c \beta_2} + 1 - \beta_2\right) \bfg^2 \\
&= \frac{\alpha (1 - \beta_2) c \beta_2 + (1 - \beta_2) (1 - c \beta_2)}{1 - c\beta_2} \cdot \bfg^2 \\
&= \frac{(\alpha - 1) (1 - \beta_2)c \beta_2 + (1 - \beta_2) }{1 - c\beta_2} \cdot \bfg^2\\
&= \frac{1 - \beta_2}{1 - c\beta_2} \cdot \left(1 +  (\alpha - 1)c\beta_2\right) \cdot \bfg^2.
\end{align*}

So, in expectation in each iteration the multiplicative error reduces from $\alpha - 1$ to $(\alpha - 1)c\beta_2$. Under our assumption $c\beta_2 < 1$ this means in expectation $\hat{\bnu}_t$ will be closer to the truth $\bfg^2$ than $\hat{\bnu}_{t-1}$. In other words, when using \cref{fig:adam-pc} not only is the ``steady state'' of $\hat{\bnu}$ proportional to $\hat{\bnu}^*$, but $\hat{\bnu}$ will converge to its steady state in expectation.

\section{Generalized Analysis from \cref{sec:regimes}}\label{sec:regimes-general}

Similarly to \cref{eq:hatnurecurrence-simple}, we have:

\[\hat{\bnu}_1 = \left(\bfg + \frac{\clipnorm}{B} \bfz_1\right)^2, \qquad
\hat{\bnu}_t = \frac{\beta_2 - \beta_2^t}{1 - \beta_2^t}\hat{\bnu}_{t-1} + \frac{1-\beta_2}{1-\beta_2^t}\left(\bfg + \frac{\clipnorm}{B} \bfz_t\right)^2.\]

Unrolling the recursion we have:

\[\hat{\bnu}_t = \sum_{j=1}^t \frac{(1-\beta_2)\beta_2^{t-j}}{1-\beta_2^t} \left(\bfg + \frac{\clipnorm}{B}\bfz_j\right)^2 = \hat{\bnu}_t^* + \sum_{j=1}^t \frac{(1-\beta_2)\beta_2^{t-j}}{1-\beta_2^t} \left(\frac{\clipnorm}{B}\bfg \bfz_j + \frac{\clipnorm^2}{B^2}\bfz_j^2\right).\]

Since $\bfz_t$ is a Gaussian and thus symmetric, $\Cov{\bfz_t}{\bfz_t^2} = 0$. So the variance of $\hat{\bnu}_t$ is

\begin{align*}
\Var{\hat{\bnu}_t} &= \left(\frac{1-\beta_2}{1-\beta_2^t}\right)^2 \sum_{j=1}^t\beta_2^{2(t-j)} \Var{\frac{\clipnorm}{B}\bfg \bfz_j + \frac{\clipnorm^2}{B^2}\bfz_j^2} \\
&= \left(\frac{1-\beta_2}{1-\beta_2^t}\right)^2 \sum_{j=1}^t\beta_2^{2(t-j)} \left(\frac{\clipnorm^2 \bfg^2}{B^2}\Var{ \bfz_j} + \frac{\clipnorm^4}{B^4}\Var{\bfz_j^2}\right)\\
&= \left(\frac{\clipnorm^2 \bfg^2 \sigma^2}{B^2} + \frac{2 \clipnorm^4 \sigma^4}{B^4}\right) \left(\frac{1-\beta_2}{1-\beta_2^t}\right)^2 \sum_{j=1}^t\beta_2^{2(t-j)}\\
&= \left(\frac{\clipnorm^2 \bfg^2 \sigma^2}{B^2} + \frac{2 \clipnorm^4 \sigma^4}{B^4}\right) \left(\frac{1-\beta_2}{1-\beta_2^t}\right)^2 \frac{1 - \beta_2^{2t}}{1 - \beta_2^2}\\
&= \left(\frac{\clipnorm^2 \bfg^2 \sigma^2}{B^2} + \frac{2 \clipnorm^4 \sigma^4}{B^4}\right) \frac{(1 - \beta_2)^2}{1 - \beta_2^2} \cdot \frac{1 - \beta_2^{2t}}{(1 - \beta_2^t)^2} \\
&\geq \left(\frac{\clipnorm^2 \bfg^2 \sigma^2}{B^2} + \frac{2 \clipnorm^4 \sigma^4}{B^4}\right) \frac{(1 - \beta_2)^2}{1 - \beta_2^2} \\
\end{align*}

In the last step we use the fact that $\beta_2^t < 1$, $\frac{1-x^2}{(1-x)^2}$ is equal to 1 at $x = 0$ and increasing in $x \in [0, 1)$. For the standard choice $\beta_2 = .999$, $\frac{(1 - \beta_2)^2}{1 - \beta_2^2}$ is equal to roughly $1 / 2000$. 

Now, in order for the bias corrected estimate $\hat{\bnu}_t - \frac{\clipnorm^2 \sigma^2}{B^2}$ to be non-negative with at least, say, probability $1/2$, a necessary condition is for the standard deviation to be within a constant factor of the magnitude of $E[\hat{\bnu}_t - \frac{\clipnorm^2 \sigma^2}{B^2}] = \bfg^2$. Rearranging, we get:

\begin{align*}
\sqrt{\Var{\hat{\bnu}_t}} &\lesssim \bfg^2 \\
\Var{\hat{\bnu}_t} &\lesssim \bfg^4\\
\rightarrow \left(\frac{\clipnorm^2 \bfg^2 \sigma^2}{B^2} + \frac{\clipnorm^4 \sigma^4}{B^4}\right) \frac{(1 - \beta_2)^2}{1 - \beta_2^2} &\lesssim \bfg^4
\\
\rightarrow B / \sigma &\gtrsim \left(\frac{(1 - \beta_2)^2}{1 - \beta_2^2}\right)^{1/4} \frac{\clipnorm}{|\bfg|}.\\
\end{align*}

In the one-dimensional case, if the clip norm $\clipnorm$ is chosen appropriately we might (optimistically) expect $\bfg \approx \clipnorm$, in which case this condition is easily satisfied even for small batch sizes (as a function of $\sigma$). Even for a more pessimistic $\bfg \approx \clipnorm / \sqrt{B}$ (what we would get if $\bfg$ was sampled to be $\clipnorm$ or $-\clipnorm$ w.p. $1/2$ each), this condition is still easily satisfied.

\mypar{The optimistic high-dimensional case} In the $\mdim$-dimensional setting, at least $1/2$ the coordinates of $\bfg_t$ (the average of the clipped gradients in round $t$) satisfy $\bfg_t(i) \le \sqrt{2} \ltwo{\bfg_t} / \sqrt{\mdim}$. So even if we optimistically assume $\ltwo{\bfg_t} = \clipnorm$ (which assumes all examples gradients are the same)
the sufficient condition for, say, at most $1/4$ of the coordinates of $\bfg_t$ having negative $\hat{\bnu}$ with probability $1/2$ becomes:

\begin{equation}\label{eq:regimes-general}
B / \sigma \gtrsim \left(\frac{(1 - \beta_2)^2}{1 - \beta_2^2}\right)^{1/4} \sqrt{\mdim}.
\end{equation}

In other words, as long as the dimension $\mdim$ is sufficiently small, a method such as bias correction will not produce negative estimates of the second moment frequently. However, if the dimension $\mdim$ is sufficiently large as a function of the batch size $B$, bias correction will frequently produce negative estimates of the second moment, which even with the stability constant can lead to improperly large per-coordinate learning rates (or a large stability constant to regularize these learning rates) and in turn undesirable learning outcomes (we will discuss this further in the next section). As an example, consider e.g. $B = 2048, \sigma = 1, \beta_2 = .999$, this condition says the dimension should be at most roughly $\mdim \lesssim 10^6$ to avoid undesirable learning outcomes. Accounting for the fact that some small constants are masked by the approximate inequality, this condition might reasonably hold for research-scale models, but not for slightly larger ones.

\mypar{The pessimistic high-dimensional case} If the per-example gradients in a batch were not all the same, but instead orthogonal, we would actually get $\ltwo{\bfg} \leq \clipnorm / \sqrt{B}$. If we instead plug in $|\bfg| = \clipnorm / \sqrt{B\mdim}$ for the average coordinate value into \eqref{eq:regimes-general}, we get the condition:

\begin{equation}\label{eq:regimes-pessimistic-general}
\sqrt{B} / \sigma \gtrsim \left(\frac{(1 - \beta_2)^2}{1 - \beta_2^2}\right)^{1/4} \sqrt{\mdim}.
\end{equation}

This is now a much more stringent condition. Consider e.g. $B = 2048, \sigma = 1, \beta_2 = .999$, this condition says the dimension should be at most roughly $\mdim \lesssim 2 \cdot 10^4$, far smaller than any modern model size (e.g., even a CNN for MNIST classification has $5 \cdot 10^5$ parameters).

\mypar{The high-dimensional case with prefix sums} If using independent moment estimation, we could instead use \dpmf, i.e. noisy prefix sums, to add correlated noise to $\hat{\bnu}_t$ instead. This might improve the regime in which unbiased estimates of $\hat{\bnu}_t$ avoid frequent negative estimates. For a noisy prefix sum of $T$ terms $\sum_{t=1}^T x_t$ with equal sensitivity, \dpmf can achieve variance scaling as $O(\log^3 T)$ as opposed to $O(T)$ achieved by independent noise. Due to the decay parameter $\beta_2$, $\hat{\bnu}_t$ is not a prefix sum of equally-weighted terms, but a decaying prefix sum $\sum_{t=1}^T \beta_2^{T-i} x_t$. Determining the variance reduction for decaying prefix sums is a question that to the best of our knowledge has not been addressed by the literature. We can instead optimistically assume \dpmf's variance is that of just privatizing the ``highest-weight'' term $x_T$ in each decaying prefix sum (e.g., in the decay-less case, this corresponds to assuming the variance is $O(1)$ instead of $O(\log^3 T)$). By almost the same derivation, this gives the following lower bound on the variance for e.g. bias-correction:

\[\Var{\hat{\bnu}_t} \geq \left(\frac{\clipnorm^2 \bfg^2 \sigma^2}{B^2} + \frac{2 \clipnorm^4 \sigma^4}{B^4}\right) (1 - \beta_2)^2.\]

In turn, using correlated noise improves the lower bound on $B / \sigma$ by a factor of $\left(\frac{1}{1-\beta_2^2}\right)^{1/4}$. For example, we get the following bound on $B / \sigma$ analogous to e.g. \cref{eq:regimes-general}:

\[ B / \sigma \gtrsim (1 - \beta_2)^{1/2} \sqrt{\mdim}.\]

For the standard $\beta_2 = .999$, we have $ \left(\frac{1}{1-\beta_2^2}\right)^{1/4} \approx 4.73$, i.e. a small constant improvement even with an optimistic assumption on the improvement in the variance. In short, this means that using correlated noise does not significantly increase the range of $B, \sigma, \mdim$ for which unbiased estimates of $\hat{\bnu}_t$ avoid bad learning outcomes.
\end{document}